\documentclass[letterpaper, 10 pt, conference, table]{ieeeconf}
\IEEEoverridecommandlockouts                              % This command is only needed if
\usepackage{cite}
\usepackage{etex}
\usepackage{multicol}
\usepackage{amsmath,amsfonts,amssymb}
\usepackage{float}
\usepackage{bm}
\usepackage{graphicx}
\usepackage{epstopdf}
\usepackage{epsfig}
\usepackage{xspace}
\usepackage{subcaption}
\usepackage{multirow}
\usepackage{hhline}
\usepackage[export]{adjustbox}
\usepackage{csvsimple}
\usepackage[dvipsnames]{xcolor}
\usepackage{caption}
\usepackage[colorinlistoftodos]{todonotes}
\newcommand{\journalVersion}[1]{}
\usepackage{xr}
\newcommand{\subparagraph}{} % temporary fix for titlesec error
\usepackage{titlesec}
\graphicspath{{./figures/}}
\usepackage{tabularx}
\newcolumntype{Y}{>{\centering\arraybackslash}X} % To make columns content centered

\usepackage{comment}
\usepackage[binary-units]{siunitx}
\usepackage{relsize}
\usepackage{ifthen}
\usepackage[colorinlistoftodos]{todonotes}

\usepackage[vlined,ruled,linesnumbered]{algorithm2e}
\usepackage{graphics} % for pdf, bitmapped graphics files
\usepackage{rotating}
\usepackage{color}
\usepackage{enumerate}
\usepackage[T1]{fontenc}
\usepackage{psfrag}
\usepackage{epsfig} % for postscript graphics files
\usepackage{booktabs}
\usepackage{graphicx,url}
\usepackage{array}
\usepackage{latexsym}
\usepackage{amsfonts}
\usepackage{amsmath}
\usepackage{amssymb}
\usepackage{xstring}
\usepackage{algpseudocode}
\usepackage{multirow}
\usepackage{xcolor}
\usepackage{prettyref}
\usepackage{flexisym}
\usepackage{bigdelim}
\usepackage{breqn} % load this last
\usepackage{listings}
\let\labelindent\relax
\usepackage{enumitem}
\usepackage{xspace}
\usepackage{bm}
\graphicspath{{./figures/}}
\usepackage{tikz}
\usetikzlibrary{matrix,calc}

\usepackage{mdwlist}

\let\itemize\undefined
\makecompactlist{itemize}{stditemize}

\newrefformat{prob}{Problem\,\ref{#1}}
\newrefformat{def}{Definition\,\ref{#1}}
\newrefformat{sec}{Section\,\ref{#1}}
\newrefformat{sub}{Section\,\ref{#1}}
\newrefformat{prop}{Proposition\,\ref{#1}}
\newrefformat{app}{Appendix\,\ref{#1}}
\newrefformat{alg}{Algorithm\,\ref{#1}}
\newrefformat{cor}{Corollary\,\ref{#1}}
\newrefformat{thm}{Theorem\,\ref{#1}}
\newrefformat{lem}{Lemma\,\ref{#1}}
\newrefformat{fig}{Fig.\,\ref{#1}}
\newrefformat{tab}{Table\,\ref{#1}}

\newcommand{\bdmath}{\begin{dmath}}
\newcommand{\edmath}{\end{dmath}}
\newcommand{\beq}{\begin{equation}}
\newcommand{\eeq}{\end{equation}}
\newcommand{\bdm}{\begin{displaymath}}
\newcommand{\edm}{\end{displaymath}}
\newcommand{\bea}{\begin{eqnarray}}
\newcommand{\eea}{\end{eqnarray}}
\newcommand{\beal}{\beq \begin{array}{ll}}
\newcommand{\eeal}{\end{array} \eeq}
\newcommand{\beas}{\begin{eqnarray*}}
\newcommand{\eeas}{\end{eqnarray*}}
\newcommand{\ba}{\begin{array}}
\newcommand{\ea}{\end{array}}
\newcommand{\bit}{\begin{itemize}}
\newcommand{\eit}{\end{itemize}}
\newcommand{\ben}{\begin{enumerate}}
\newcommand{\een}{\end{enumerate}}

\newcommand{\eg}{\emph{e.g.,}\xspace}
\newcommand{\ie}{\emph{i.e.,}\xspace}

\renewcommand{\boldsymbol}[1]{{\bm #1}}
\newcommand{\hide}[1]{}

\newcommand{\hiddenText}{{\color{gray} hidden text.}}
\newcommand{\hideWithText}[1]{\hiddenText}

\DeclareMathOperator*{\argmax}{arg\,max}

\newcommand{\blue}[1]{{\color{blue}#1}}

\newcommand{\linkToPdf}[1]{\href{#1}{\blue{(pdf)}}}
\newcommand{\linkToPpt}[1]{\href{#1}{\blue{(ppt)}}}
\newcommand{\linkToCode}[1]{\href{#1}{\blue{(code)}}}
\newcommand{\linkToWeb}[1]{\href{#1}{\blue{(web)}}}
\newcommand{\linkToVideo}[1]{\href{#1}{\blue{(video)}}}
\newcommand{\linkToMedia}[1]{\href{#1}{\blue{(media)}}}
\newcommand{\award}[1]{\xspace} % {{\red{#1}}} % omit awards

\newcommand{\methodname}{Hydra++\xspace}

\newcommand{\classtype}[1]{{\smaller \sf#1}}
\newcommand{\chair}{\classtype{Chair}\xspace}

\newcommand{\bin}{\classtype{Bin}\xspace}

\newcommand{\bmat}{\left[ \begin{array}}
\newcommand{\emat}{\end{array} \right]}

\newcommand{\bal}{\begin{align}}
\newcommand{\eal}{\end{align}}

\newcommand{\TSDF}{TSDF\xspace}

\newcommand{\CRISP}{CRISP\xspace}

\newcommand{\RMCC}{RMCC\xspace}
\newcommand{\RMCCfull}{reprojection-mask consistency check\xspace}
\newcommand{\dinov}{DINOv2\xspace}
\newcommand{\NOCS}{NOCS\xspace}
\newcommand{\NOCSfull}{normalized object coordinate space\xspace}
\newcommand{\activewindow}{active window\xspace}

\newcommand{\IoU}{\mathrm{IoU}\xspace}
\newcommand{\RGBD}{RGB-D\xspace}

\newcommand{\LiDAR}{LiDAR\xspace}

\newcommand{\SDF}{SDF\xspace}

\newcommand{\imgRGB}{\mathbf{I}_{\mathrm{RGB}}}          % RGB image
\newcommand{\imgDepth}{\mathbf{I}_\mathrm{depth}}        % Depth image
\newcommand{\imgInst}{\mathbf{I}_{\mathrm{inst}}}         % Instance segmentation image
\newcommand{\vertexMap}{\mathbf{V}}                       % Vertex map (back-projected 3D)

\newcommand{\camMatrix}{\mathbf{K}}                    % Camera intrinsic matrix
\newcommand{\pixelCoord}{\mathbf{u}}                   % Projected pixel coordinates

\newcommand{\maskObs}{\mathbf{M}_{\mathrm{obs}}}         % Observed mask
\newcommand{\maskProj}{\mathbf{M}_{\mathrm{proj}}}       % Projected mask
\newcommand{\posvec}{\mathbf{p}}                       % Position vector (generic)
\newcommand{\rotmat}{\mathbf{R}}                       % Rotation matrix
\newcommand{\tfmat}{\mathbf{T}}                        % Transformation matrix (SE(3))

\newcommand{\shapecode}{\mathbf{z}_{\mathrm{shape}}}     % Shape latent code
\newcommand{\inputvec}{\mathbf{x}}                     % Input vector
\newcommand{\nodeSet}{\mathcal{V}}                     % Node/vertex set
\newcommand{\edgeSet}{\mathcal{E}}                     % Edge set
\newcommand{\subPred}{\mathrm{pred}}                    % Predicted subscript
\newcommand{\subGT}{\mathrm{GT}}                        % Ground-truth subscript

\newcommand{\pointSet}{\mathcal{P}}                    % Point set
\newcommand{\pointSetPred}{\mathcal{P}_{\subPred}}
\newcommand{\pointSetGT}{\mathcal{P}_{\subGT}}    
\newcommand{\cluster}{\mathcal{C}}                  % Cluster set
\newcommand{\voxelSet}{\mathcal{U}}                    % Voxel set
\newcommand{\voxelElem}{\boldsymbol{u}}                    % Voxel set

\newcommand{\worldFrame}{{}^W}                         % World frame superscript
\newcommand{\objectcoord}{O}

\newcommand{\worldTcamt}{{}^W\mathbf{T}_{C_{t^*}}}              % World to camera transform
\newcommand{\camTobj}{{}^C\mathbf{T}_O}                         % Camera to object transform (generic)

\newcommand{\voxelMap}{\mathcal{M}}                    % Volumetric map
\newcommand{\tsdfval}{\tau}     % TSDF layer
\newcommand{\objMesh}{\mathcal{M}}                       % Object mesh (general representation)
\newcommand{\dsg}{\mathcal{G}}                         % Dynamic scene graph

\newcommand{\track}{\mathcal{T}}                       % Individual track
\newcommand{\trackIndex}{r}                       % Individual track
\newcommand{\surfIndex}{j}                       % Surface point index

\DeclareMathOperator{\Volume}{vol}                     % Volume function
\DeclareMathOperator{\BBox}{3DBBox}                      % Bounding box function
\newcommand{\precisionVar}{O}                          % Precision metric variable
\usepackage[colorlinks=true, allcolors=blue, bookmarks=true]{hyperref}

\usepackage{mathtools}

\usepackage[capitalize]{cleveref}

\Crefname{section}{Sec.}{Secs.} 
\Crefname{figure}{Fig.}{Figs.} 
\Crefname{table}{Table}{Tables} 
\crefname{equation}{}{}
\title{\huge{\methodname: Real-Time Hierarchical 3D Scene Graph Construction With Object-Level Shape Estimation}}

\author{Hyungtae Lim$^{1}$, Nathan Hughes$^{1}$, Xihang Yu$^{1}$, Ruihan Xu$^{1}$, \\ Yun Chang$^{1}$, Jingnan Shi$^{1}$, Rajat Talak$^{2}$, and Luca Carlone$^{1}$
\thanks{$^{1}$H. Lim, N. Hughes, X. Yu, R. Xu, Y. Chang, J. Shi, and L. Carlone are with the Laboratory for 
Information \& Decision Systems, Massachusetts Institute of Technology, Cambridge, MA, USA, 
{\sf \{shapelim, na26933, jimmyyu, multyxu, yunchang, jnshi, lcarlone\}@mit.edu}
}
\thanks{$^{2}$R. Talak is with the Department of Electrical and Computer Engineering at the National University of Singapore, Singapore,  
{\sf \{talak@nus.edu.sg\}}
}
\thanks{This work was
partially funded by 
the National Research Foundation of Korea (NRF) grant funded by the Korean government (MSIT) (No. RS-2024-00461409) and by the Department of the Air Force Artificial Intelligence Accelerator, accomplished under Cooperative Agreement Number FA8750-19-2-1000.
}
}

\begin{document}

\makeatletter
  \let\@oldmaketitle\@maketitle% Store \@maketitle
  \renewcommand{\@maketitle}{\@oldmaketitle% Update \@maketitle to insert...
  \bigskip
  \centering
    % This `\setcounter{figure}{0}` should be here
    \setcounter{figure}{0}
    \includegraphics[width=0.9\textwidth]{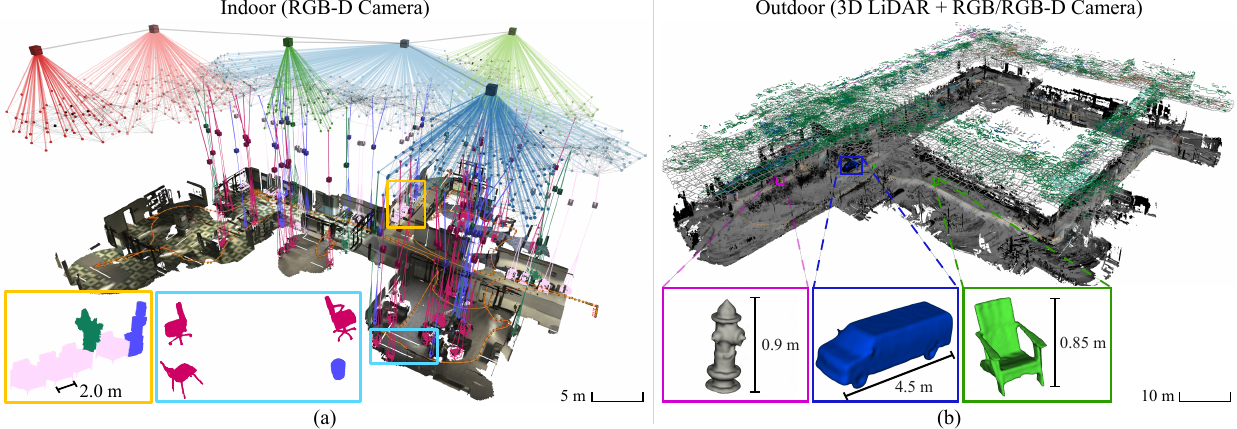}
    \vspace{-1mm}
    \captionof{figure}{3D scene graphs with object-level shape reconstruction generated by \textit{\methodname}.
      (a)~Indoor environment using the uHumans2 dataset~\cite{Rosinol21ijrr-Kimera}.
      Zoom-in views show full object shapes reconstructed from partial observations. Each color represents a semantic class~(\eg~\classtype{couches} in pink, \classtype{plants} in green, \classtype{trash bins} in purple, and \classtype{chairs} in red).
      The object shape estimation network in our framework captures intra-class variation,
      as illustrated by the reconstructed \classtype{trash bins} and \classtype{chairs} with different geometries.
      (b)~Outdoor campus scene using the hybrid \LiDAR-camera configuration.
      Our method reconstructs objects of various sizes,
      from a small \classtype{fire hydrant} to a large \classtype{car}, under realistic sensing conditions.}
    \label{fig:fig1}
    \vspace{-3mm}
  }%an image
\makeatother

\maketitle

% \bstctlcite{bst_control}

%!TEX root = ../main.tex

\begin{abstract}
3D scene graphs provide a hierarchical abstraction of environments by encoding spatial entities (\eg objects, places) and their relationships.
However, existing scene graph systems model object geometry coarsely, relying on partial point clouds or class-level CAD templates, which limits instance-specific shape detail.
This paper presents \textit{\methodname}, a system-level investigation
into how learning-based object shape estimators can be integrated into a hierarchical 3D scene graph pipeline.
\methodname incorporates category-agnostic shape estimation and a reprojection-mask consistency check~(\RMCC) to reject degenerate predictions from partial observations or imprecise segmentation.
In its default \CRISP-based configuration, \methodname performs online scene graph construction; slower estimators such as SAM3D are evaluated as modular alternatives to demonstrate generalization-latency trade-offs.
Furthermore, to address the challenges of sparse and noisy depth measurements in outdoor environments,
\methodname supports a hybrid \LiDAR-camera configuration for large-scale operation,
improving scene-level reconstruction quality.
Experiments in both simulation and real-world outdoor campus scenarios demonstrate that \methodname improves object- and scene-level reconstruction quality.
Project page is available at \url{https://hydra-plusplus.github.io/}.
\end{abstract}

%!TEX root = ../main.tex
\section{Introduction}
\label{sec:introduction}

Modeling 3D scenes is crucial for many robotics tasks, from navigation to manipulation~\cite{Hughes22rss-hydra}.
With the rise of deep learning and large language models (LLMs), 
many studies have examined semantic mapping, which captures semantic information about the environment to support reasoning and interaction~\cite{Salas-Moreno13cvpr,Nicholson18ral-quadricSLAM,Yang19tro-CubeSLAM,Li16iros-metricSemantic}.
In particular, recent 3D scene graph approaches encode semantics by treating entities~(\eg objects, rooms, and agents) as nodes,
and relationships as edges (\eg adjacency, inclusion, and support), 
yielding a lightweight and efficient abstraction of the environment~\cite{Armeni19iccv-3DsceneGraphs,Hughes24ijrr-hydraFoundations}.

However, most scene graph systems still treat object geometry as a coarse proxy, typically relying on centroids and bounding boxes
with partial reconstructions from accumulated point clouds~\cite{Hughes22rss-hydra,Wu21cvpr-SceneGraphFusion}.
This abstraction is often sufficient for semantic queries~\cite{Honerkamp24ral-LanguageGroundede},
but it becomes a bottleneck when downstream tasks require instance-specific geometry, such as contact-aware reasoning, model-based manipulation, or rearrangement.
In outdoor environments, this limitation compounds with sparse and noisy depth measurements,
making mesh reconstruction of both objects and background especially challenging.
Together, these challenges motivate the need for higher-fidelity object and background reconstructions, especially in large-scale outdoor environments.

% \begin{figure}[t!]
%  	\centering
%  	\begin{subfigure}[b]{0.45\textwidth}
%  		\includegraphics[width=1.0\textwidth]{figures/tbu.pdf}
%  	\end{subfigure}
%  	\captionsetup{font=small}
%   \caption{TBU.}
%     \label{fig:fig1}
%     % \vspace{-6mm}
% \end{figure}

In this paper, we present \textit{\methodname}, a system-level study of how object shape estimators can be integrated into a hierarchical 3D scene graph pipeline.
As shown in \Cref{fig:fig1},
\methodname extends the frameworks of Hydra~\cite{Hughes24ijrr-hydraFoundations} and Khronos~\cite{Schmid24rss-khronos} 
with per-object mesh-level reconstruction by incorporating category-agnostic shape estimators (\CRISP~\cite{Shi25cvpr-CRISP} or SAM3D~\cite{Chen2026CVPR-sam3d}).
To handle degenerate shape predictions arising from partial observations or imprecise segmentation, we introduce a reprojection-mask consistency check (\RMCC).
We further study a hybrid \LiDAR-camera configuration that combines the metric-scale depth from a 3D \LiDAR sensor with image-based shape estimation, improving reconstruction quality at both the object level and the scene level in outdoor settings.
Throughout the paper, the real-time claim refers to the default \CRISP-based instantiation of \methodname.
SAM3D is included as an alternative module to analyze the trade-off between out-of-domain generalization and inference latency.

% In sum, we provide three key contributions.
% We (i)~develop a system to integrate learning-based object shape estimation into hierarchical 3D scene graph construction,
% maintaining robustness to partial observations and segmentation noise;
% (ii)~analyze the use of different object pose and shape estimators, demonstrating that the pipeline remains model-agnostic and exposing practical trade-offs between model choices; and
% (iii)~extend existing scene graph pipelines to fuse \LiDAR and camera information, improving both scene-level mesh continuity and object-level geometric reconstruction in outdoor environments.
% These contributions are substantiated by the corresponding experimental sections.

With \methodname, we claim the following contributions.
We (i)~introduce an enhanced 3D scene graph construction system that integrates learning-based object shape estimation models
to provide high-fidelity object geometry while remaining robust to noisy object detections,
(ii)~present a study of trade-offs between state-of-the-art shape-estimation methods, and
(iii)~extend the system with a hybrid \LiDAR-camera configuration to support large-scale outdoor environments.
We support these claims with experiments in both simulated and real-world environments.

%!TEX root = ../main.tex

\newcommand{\etalcite}[1]{\textit{et al.}~\cite{#1}}

\section{Related Work}
\label{sec:related_work}

\noindent{\bf Object-level SLAM.}
3D scene understanding is a fundamental task in robotic mapping,
with extensive research enabling robots to understand and represent environments at varying semantic granularities. %~\cite{Salas-Moreno13cvpr,Dong17cvpr-XVIO,Mo19iros-orcVIO,Nicholson18ral-quadricSLAM,Bowman17icra,Ok21icra-home,Wu21cvpr-SceneGraphFusion}.
Pioneered by SLAM++~\cite{Salas-Moreno13cvpr}, object-level SLAM has since incorporated semantic objects directly into the mapping and localization pipeline~\cite{Nicholson18ral-quadricSLAM,Yang19tro-CubeSLAM,Li16iros-metricSemantic}.
Representative systems such as QuadricSLAM~\cite{Nicholson18ral-quadricSLAM} and CubeSLAM~\cite{Yang19tro-CubeSLAM} model objects using parametric geometric primitives within factor graph formulations, enabling lightweight and interpretable object-level mapping.
While effective, these approaches generally rely on predefined geometric models,
which limits their ability to adapt to partially observed or irregular object geometries.

To address this limitation, subsequent works extend this line of
research through robust object data association and optimization.
Bowman~\etalcite{Bowman17icra} tackle the data association problem
directly by formulating a joint optimization over metric and semantic
information with probabilistic object associations, enabling reliable
object-level mapping in ambiguous environments.
Building on this line of work, Atanasov~\etalcite{Atanasov18ijcai-Unifying}
provide a unifying view of geometry, semantics, and data association
in SLAM, introducing structured object models of mid-level part features
to generalize beyond purely geometric representations.

\vspace{2mm}
\noindent{\bf Learning-Based Object Shape Representations.}
Further advancing the expressiveness of object representations,
Shan~\etalcite{Shan21iccv-ELLIPSDF} propose \textit{ELLIPSDF}, which jointly
optimizes object pose and shape using a bi-level model consisting of
a coarse ellipsoid and a fine-grained neural SDF, enabling compact yet
expressive object-level map inference from multi-view RGB-D observations.
Shape estimation for objects has since been explored through a variety
of paradigms, including parametric fitting~\cite{Yu2025box},
volumetric representations~\cite{Xu19icra-MID-Fusion},
learning-based implicit functions~\cite{Park19cvpr-deepSDF,Shi25cvpr-CRISP},
and diffusion-based approaches~\cite{Liu24neuips-oneTwoThree},
collectively demonstrating the feasibility of generating detailed object
geometries even from partial observations.

In parallel, the integration of shape priors for detailed object
representations has been explored~\cite{Wang213dv-DSP-SLAM},
while decentralized and collaborative formulations have been proposed
to support multi-robot object-level mapping under communication
constraints, exemplified by SlideSLAM~\cite{Liu25tro-SlideSLAM}.
% More recently, SAM3D~\cite{Chen2026CVPR-sam3d} shows promising category-agnostic
% object shape reconstruction without task-specific training
% and Siddiqui~\etalcite{Siddiqui2026ShapeR} highlights the growing potential of 
% integrating detailed object-level shape estimation directly into mapping pipelines.

\vspace{2mm}
\noindent{\bf Scene Representations With 3D Scene Graphs.}
These developments in object-level representation collectively point
toward a growing interest in organizing rich object geometry and
semantics within structured spatial frameworks.
Scene graph-based methodologies have emerged as a promising paradigm
for such hierarchical spatial
understanding~\cite{Armeni19iccv-3DsceneGraphs,Rosinol21ijrr-Kimera,
Rosinol20rss-dynamicSceneGraphs}, enabling robots to reason about
environments from low-level geometric primitives to high-level semantic
concepts such as rooms and buildings.
Hughes~\etalcite{Hughes24ijrr-hydraFoundations} present Hydra,
a real-time spatial perception system that incrementally constructs
layered scene graphs, handling loop closures through hierarchical
descriptors and deformation-based optimization.
Bavle~\etalcite{Bavle22ral-SGraph,Bavle22arxiv-SGraphPlus} also
develop \LiDAR-based hierarchical representations for indoor scenes,
called \textit{S-Graphs} and \textit{S-Graphs+}.
Werby~\etalcite{Werby23rss-HOVSG} extend this paradigm by
incorporating open-vocabulary vision-language features,
enabling language-conditioned queries in multi-story environments.
Greve~\etalcite{Greve24icra-CURBSG} further push scene graphs toward outdoor autonomy,
constructing collaborative dynamic scene graphs from multi-agent panoptic \LiDAR data
for large-scale urban environments.

Although object-level SLAM and scene graph representations are each
advancing rapidly, there remains limited exploration of how detailed
learned object shapes can be organized within hierarchical scene
understanding frameworks.
Existing scene graph systems excel at capturing semantic and topological
relationships but typically abstract object geometry to simple
primitives, while shape estimation methods produce detailed object
models but often operate in isolation from broader spatial context.
In this work, we investigate the integration of learned object mesh
estimation within scene graph representations
using \CRISP~\cite{Shi25cvpr-CRISP} or SAM3D~\cite{Chen2026CVPR-sam3d},
and explore how such
representations can be incorporated into hierarchical spatial
understanding pipelines to enable applications that benefit from both
semantic awareness and object-level geometric detail.

% ===== In your document =====
\section{Methodology}
\label{sec:methodology}

\subsection{System Overview}\label{sec:overview}

\methodname builds on the Hydra~\cite{Hughes24ijrr-hydraFoundations} and Khronos~\cite{Schmid24rss-khronos} frameworks for 3D scene graph construction, 
extending them with a learning-based shape estimation module~\cite{Shi25cvpr-CRISP,Chen2026CVPR-sam3d}, a geometric consistency check for predicted object shapes,
and support for a hybrid \LiDAR-camera sensor configuration.

\begin{figure}[!t]
    \centering
    \includegraphics[trim={0 0 0 0},clip, width=0.95\columnwidth]{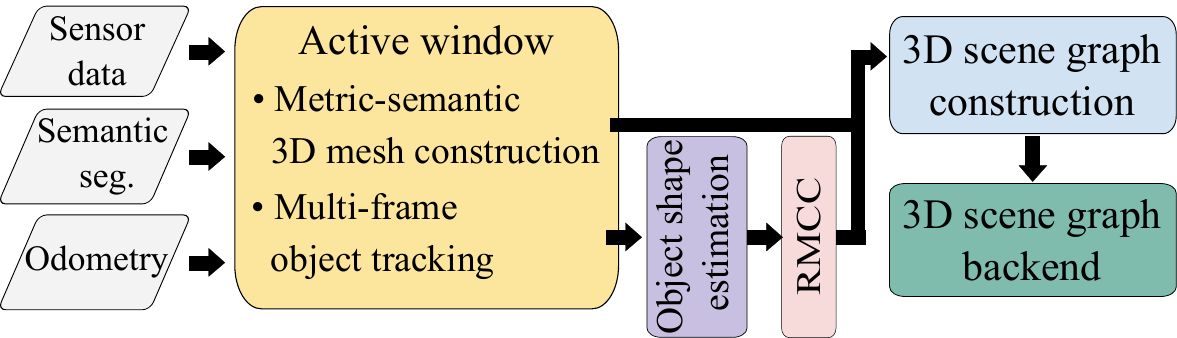}
    \caption{System pipeline of \textit{\methodname}. Sensor data (\RGBD or 3D \LiDAR + camera, semantic segmentation, and odometry) flows through an \activewindow module that performs metric-semantic 3D mesh construction and object tracking.
    Inactive tracks are processed by a learning-based shape estimation module (instantiated as \CRISP~\cite{Shi25cvpr-CRISP} or SAM3D~\cite{Chen2026CVPR-sam3d}) for object pose and shape estimation,
    which is followed by our proposed \RMCCfull (\RMCC) to validate the predictions.
    The 3D scene graph construction module instantiates nodes across multiple semantic layers, which are then optimized and updated asynchronously by the backend through deformation-graph-based optimization and incremental updates~\cite{Rosinol21ijrr-Kimera}.}
    \label{fig:diagram}
    \vspace{-2mm}
\end{figure}

As illustrated in \Cref{fig:diagram}, the system receives sensor data (\RGBD or \LiDAR + RGB/\RGBD camera), instance segmentation masks, and odometry estimates.
From these inputs, an \emph{\activewindow} module maintains a volumetric map~$\voxelMap_t$ that is incrementally updated via projective truncated signed distance field~(\TSDF) integration~\cite{Schmid22icra-Multi-tsdf}.
% Regions occupied by moving objects are detected by comparing observed and ray-cast depths, and a binary integration mask excludes them from \TSDF integration to limit corruption of the static metric-semantic mesh.
Meanwhile, object candidates are extracted in image space from the instance label image~$\imgInst$,
which encodes per-pixel category and instance IDs from an off-the-shelf segmentation model.
Pixels sharing the same instance ID are grouped into clusters~$\cluster_i^t$, 
each lifted to 3D by back-projecting the depth image through the camera intrinsics and depth information into a vertex map~$\vertexMap$.
Clusters are filtered by sensing range, pixel count, and 3D bounding-box volume.
A max-$\IoU$ tracker~\cite{Schmid24rss-khronos} then associates clusters across frames using voxel-based overlap,
forming temporal object tracks~$\track_k$, where $k$ indexes the currently active tracks.

When a track becomes inactive (\ie the object moves outside the active window), 
the system invokes a learning-based shape estimation module for 6-DoF pose and shape estimation (\Cref{sec:shape_est}),
followed by a \RMCCfull~(\RMCC) to validate the prediction (\Cref{sec:rmcc}).
Validated object nodes, together with place and region nodes,
are assembled into a hierarchical 3D scene graph~$\dsg=(\nodeSet,\edgeSet)$ using the Hydra backend~\cite{Hughes24ijrr-hydraFoundations},
where nodes can span multiple semantic layers (\eg agents, objects, places, and regions) and edges encode spatial and semantic relationships.

In the remainder of this section, we describe the two primary contributions of \methodname:
(a)~an adaptive integration strategy for improved ground reconstruction in the hybrid \LiDAR-camera configuration (\Cref{sec:adaptive}), and
(b)~the object-level pipeline comprising a learning-based shape estimation module (\Cref{sec:shape_est}) and a geometric consistency check via \RMCC (\Cref{sec:rmcc}).

% \subsection{Overview of Our System}

\subsection{Ground-Aware Adaptive Integration Strategy}\label{sec:adaptive}

\begin{figure}[!t]
    \centering
  	\begin{subfigure}{0.48\columnwidth}
      \includegraphics[trim={0 0 0 0},clip, width=\columnwidth]{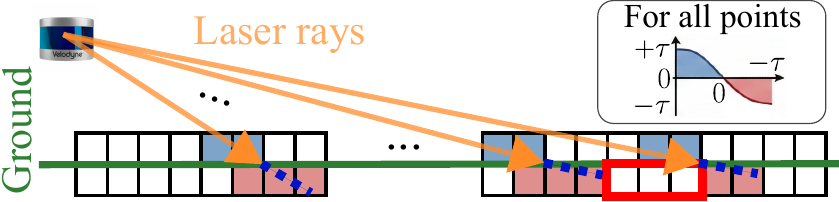}
      \includegraphics[trim={0 0 0 0},clip, width=\columnwidth]{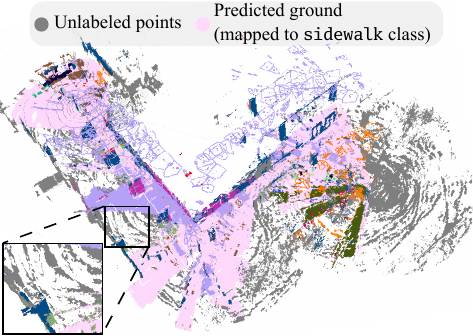}
      \caption{}
   	\end{subfigure}
  	\begin{subfigure}{0.48\columnwidth}
      \includegraphics[trim={0 0 0 0},clip, width=\columnwidth]{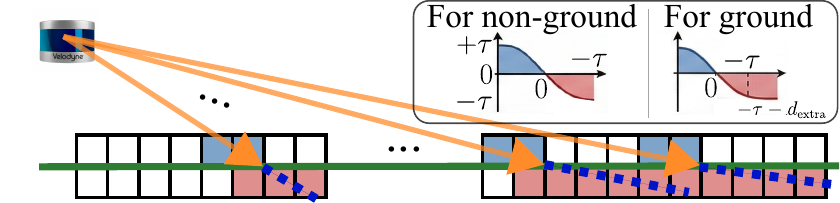}
      \includegraphics[trim={0 0 0 0},clip, width=\columnwidth]{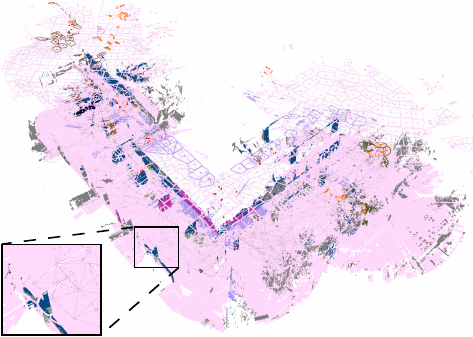}
      \caption{}
   	\end{subfigure}
    \caption{Comparison of \TSDF integration behavior and resulting ground mesh quality.
    Top: \LiDAR laser rays hit the ground at high incidence angles, especially at long range.
    (a)~Since standard integration drops measurements with signed distances beyond the truncation boundary~$-\tau$, 
    at long range the oblique ray overestimates the signed distance and the negative-side voxels receive no updates~(white voxels in the red box),
    breaking the sign transitions.
    (b)~Our ground-aware adaptive integration strategy allows measurements within an additional range $d_{\text{extra}}$ beyond $-\tau$
    for ground-labeled voxels while keeping the standard band for non-ground voxels,
    ensuring that negative-side voxels are updated even at large incidence angles.
    Bottom: the resulting mesh quality, where (a)~standard integration produces fragmented ground geometry, while (b)~our method recovers a complete and consistent ground mesh.}
    \label{fig:adaptive}
\end{figure}

In \TSDF-based mapping, each voxel $\voxelElem$ within the active window stores a signed distance value $\phi(\voxelElem)$:
positive in free space~(\ie~in front of the surface), negative behind a surface, and zero at the surface itself.
Surface meshes are extracted by finding the zero level-set $\{\voxelElem\mid\phi(\voxelElem)=0\}$,
which in practice reduces to detecting sign changes between neighboring voxels via marching cubes.
For a sign change to exist, voxels on \textit{both sides} of a surface must receive integration updates; 
however, voxels beyond the truncation distance $\tsdfval$ behind a surface are never updated and carry no weight.
This means that if the negative-side voxels of a surface are not reached by any measurement,
no sign change is recorded and the surface simply goes missing~(\ie~red box in \Cref{fig:adaptive}(a)).

In projective integration, the signed distance is estimated along the ray rather than the surface normal~\cite{Schmid24rss-khronos},
which overestimates the true distance at high incidence angles and pushes measurements outside the truncation band,
leaving negative-side voxels without updates.
% In \RGBD pipelines this is rarely an issue:
% cameras observe surfaces at moderate incidence angles with dense depth coverage,
% so negative-side voxels are reliably updated and sign changes are consistently captured.
% However, this assumption breaks down in the hybrid \LiDAR-camera mode.
When using \LiDAR point clouds, this effect is compounded by the sparsity of \LiDAR returns on the ground plane at long range,
causing inconsistent sign transitions and fragmented or missing ground meshes.

To address this, we introduce a ground-aware adaptive integration strategy.
For ground-labeled voxels~\cite{Lim21ral-Patchwork}, instead of discarding measurements whose signed distance
falls below the truncation distance $-\tsdfval$, 
we allow measurements whose signed distance lies within 
an additional range $d_{\text{extra}}$ beyond the truncation band~(empirically, $d_{\text{extra}} = 3$\,m), as illustrated by the extended dotted lines in \Cref{fig:adaptive}(b).
Measurements falling within this extended band are integrated by clamping their observed distance
to $-\tsdfval$ with a small constant weight, rather than being discarded.
This controlled negative support ensures that voxels behind the ground surface receive integration updates,
restoring the sign transitions required for reliable zero level-set extraction.

\subsection{Object Pose and Shape Estimation}\label{sec:shape_est}

As mentioned in \Cref{sec:overview}, when an object track $\track_k$ becomes inactive,
the system invokes a learning-based pose and shape estimation module to predict the 6-DoF object pose and geometry
from the observation in which the object is most fully visible.
To identify this observation, we heuristically select the frame with the maximum 3D bounding-box volume, which typically corresponds to the view with minimal occlusion:
\begin{equation}
  (\cluster^*,t^*)=\argmax_{(\cluster^{t_\trackIndex},t_\trackIndex)\in\track_k}\ \Volume\!\left(\BBox(\cluster^{t_\trackIndex})\right).
  \label{eq:selected}
\end{equation}
For the selected observation, we extract the \RGBD frame and binary mask $\maskObs\in\{0,1\}^{H\times W}$ to form the module input:
\begin{equation}
\inputvec=\left(\imgRGB,\,\imgDepth,\,\maskObs,\,\camMatrix\right),
\label{eq:shape_input}
\end{equation}
where $\imgRGB\in\mathbb{R}^{H\times W\times 3}$ is the RGB image, $\imgDepth\in\mathbb{R}^{H\times W}$ is the depth image obtained from either an \RGBD camera or a \LiDAR point cloud projected onto the image plane, and $\camMatrix\in\mathbb{R}^{3\times 3}$ is the camera intrinsic matrix.
The module outputs an object pose $\camTobj\in\mathrm{SE}(3)$, a metric scale $s\in\mathbb{R}_+$, and an object mesh $\objMesh$.
The object pose in the world frame~$\worldFrame\tfmat_{\objectcoord}$ is:
\begin{equation}
\worldFrame\tfmat_{\objectcoord}=\worldTcamt \cdot \camTobj,
\end{equation}
where $\worldTcamt$ is the camera pose at time $t^*$ in \Cref{eq:selected}.
Each object node stores:
\begin{equation}
\mathcal{A}_{\text{obj}} = \{l,\ s,\ \objMesh,\ \worldFrame\posvec_{\objectcoord},\ \worldFrame\rotmat_{\objectcoord}\},
\label{eq:object_attributes}
\end{equation}
where $l$ is the semantic label, $s\in\mathbb{R}_+$ is the metric scale, $\objMesh$ is the object mesh, and $\worldFrame\posvec_{\objectcoord}\in\mathbb{R}^3$ and $\worldFrame\rotmat_{\objectcoord}\in\mathrm{SO}(3)$ are the translation and rotation of $\worldFrame\tfmat_{\objectcoord}$, respectively.

Next, we describe the integration with two different pose and shape estimators.
Note that our framework is modular and agnostic to the underlying shape estimation architecture, making it readily extensible to other methods.

\noindent \textbf{CRISP:}\, \CRISP~\cite{Shi25cvpr-CRISP} is a category-agnostic network that jointly estimates object mesh shape and 6-DoF pose and metric scale via \NOCSfull (\NOCS) prediction.
It employs a \dinov backbone~\cite{Oquab23arxiv-dinov2} with parallel branches for shape reconstruction and \NOCS prediction, respectively, yielding:
\begin{equation}
\camTobj\in\mathrm{SE}(3),\quad s\in\mathbb{R}_{+},\quad \shapecode\in\mathbb{R}^{256},
\label{eq:crisp_output}
\end{equation}
where $\camTobj$ is the object pose relative to the \NOCS canonical frame, $s$ is the metric scale, and $\shapecode$ is the latent shape code.
Because \CRISP solves for rotation, translation, and scale jointly through pixel-wise correspondences between 3D points back-projected from $\imgDepth$ and their predicted \NOCS coordinates,
all three quantities are recovered in a single forward pass without additional processing.
The object mesh $\objMesh$ is then obtained by decoding $\shapecode$ through an implicit \SDF decoder,
extracting the iso-surface via marching cubes and scaling by $s$.

\noindent \textbf{SAM3D:}\, SAM3D~\cite{Chen2026CVPR-sam3d} outputs $\objMesh$ directly as a dense mesh along with an object pose estimate, but the mesh is defined only up to an unknown scale factor.
To recover metric scale, following the pixel-wise scale alignment used in VGGT-SLAM~\cite{Maggio25neurips-VGGT-SLAM},
we establish pixel-wise correspondences between the depth rendered by rasterizing $\objMesh$ from the camera viewpoint 
and the sensor-measured depth $\imgDepth$ over the object mask $\maskObs$,
and estimate $s$ as the median depth ratio across all valid correspondences.
% Specifically, let $\hat{d}(\mathbf{p})$ denote the depth obtained by rasterizing the predicted mesh from the camera viewpoint at pixel $\mathbf{p}$, and let $\imgDepth(\mathbf{p})$ denote the sensor-measured depth.
% The metric scale is then estimated as:
% \begin{equation}
% s = \operatorname{median}_{\mathbf{p}\,\in\,\maskObs}\left\{\frac{\imgDepth(\mathbf{p})}{\hat{d}(\mathbf{p})}\right\},
% \label{eq:sam3d_scale}
% \end{equation}
% where the median operator provides robustness to occluded or invalid depth measurements.

\subsection{Reprojection-Mask Consistency Check (RMCC)}\label{sec:rmcc}

\begin{figure}[!t]
    \centering
    \begin{subfigure}{0.48\columnwidth}
      \includegraphics[trim={0 0 0 0},clip, width=\columnwidth]{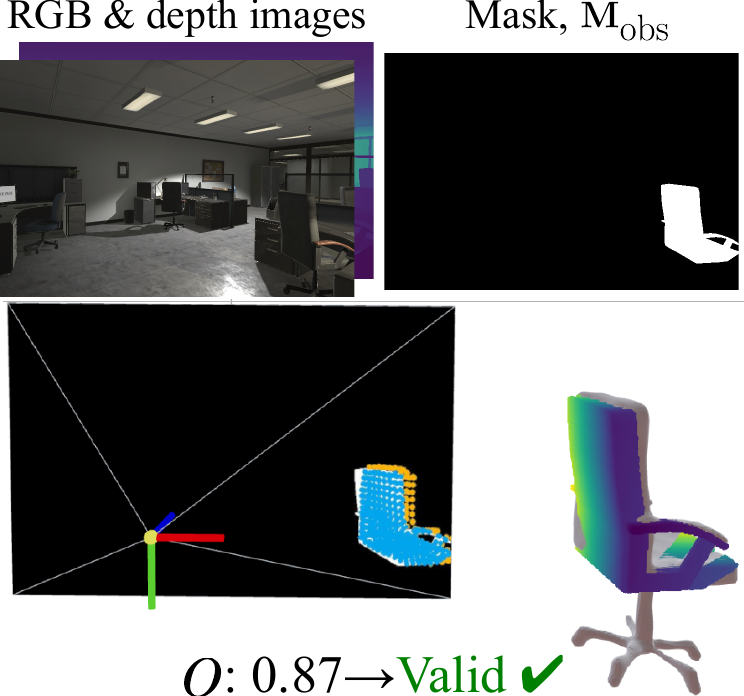}
      % \vspace{0.2em}
      % \caption{}
    \end{subfigure}
    {\color{gray!60}\rule{0.3pt}{0.4\linewidth}}
  	\begin{subfigure}{0.48\columnwidth}
      \includegraphics[trim={0 0 0 0},clip, width=\columnwidth]{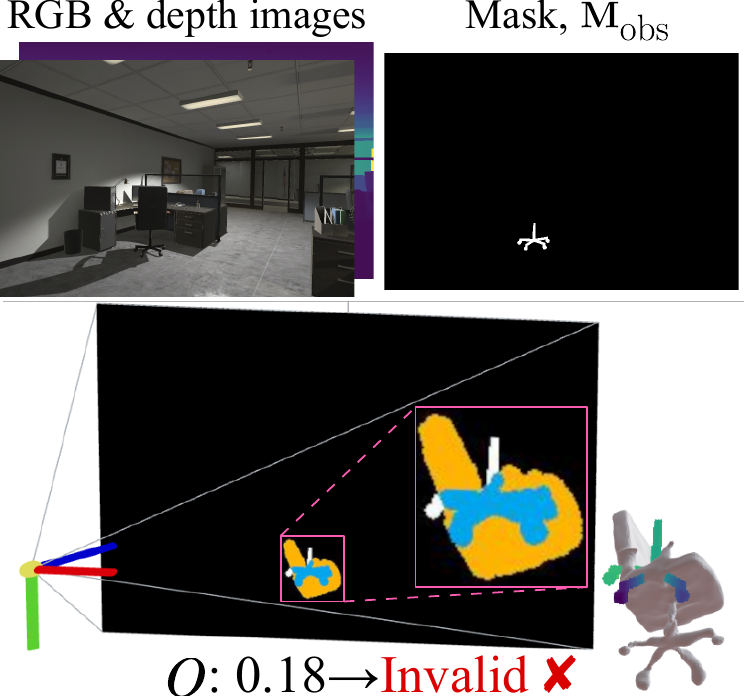}
      % \vspace{0.2em}
      % \caption{}
    \end{subfigure}
    \includegraphics[trim={0 0 0 0},clip, width=1.0\columnwidth]{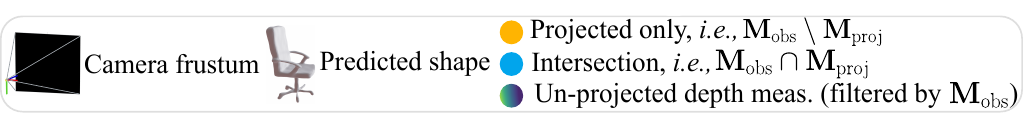}
    \caption{Overview of the reprojection-mask consistency check (RMCC) for valid (left) and invalid (right) cases.
    From top to bottom: the RGB-D image with the observed object mask $\maskObs$ (input to the shape estimation module, \Cref{sec:shape_est});
    and the camera frustum visualizing the observed mask $\maskObs$ alongside the reprojected mask $\maskProj$ generated from a point cloud sampled from the predicted shape.
    The validity of the predicted object shape is determined by the \emph{overlap score}~$\precisionVar$ between $\maskProj$ and $\maskObs$; see \Cref{eq:overlapscore}.} 
    \label{fig:rmcc}
\end{figure}

The shape estimation module can produce unreliable predictions from noisy or incomplete masks due to over-segmentation or mask imprecision.
To filter out such degenerate predictions, we employ a \RMCCfull (\RMCC) that verifies geometric alignment between the predicted shape and the observed mask (\Cref{fig:rmcc}).
\RMCC samples surface points $\pointSet_{\text{surf}}$ from the predicted mesh $\objMesh$, scales them by $s$, transforms them via $\camTobj$, and projects them onto the image plane:
\begin{align}
\pixelCoord_\surfIndex &= \pi\left(\posvec_{\text{cam},\surfIndex}; \camMatrix\right), \\
\text{where}\ \pi([x,y,z]^{\top}; \camMatrix) &= \left[\frac{f_x x}{z}+c_x,\ \frac{f_y y}{z}+c_y\right]^{\top}.
\end{align}
The projected coordinates form a mask $\maskProj\in\{0,1\}^{H\times W}$, and we compute the overlap score:
\begin{equation}
\precisionVar = \frac{|\maskProj \cap \maskObs|}{|\maskProj|}.
\label{eq:overlapscore}
\end{equation}
A prediction is accepted if $\precisionVar > \tau_\precisionVar$, filtering pose ambiguities from occluded or truncated viewpoints.

%!TEX root = ../main.tex
\section{Experiments}
\label{sec:experiments}

We conduct experiments to substantiate the three key contributions.
We (i)~evaluate the integration of learning-based object shape estimation into a hierarchical 3D scene graph pipeline~(Secs.~\ref{sec:sg_quality} and \ref{sec:intraclass});
(ii)~analyze the practical trade-offs between shape estimator choices by comparing \CRISP, our default estimator,
with SAM3D as a slower alternative instantiation~(\Cref{sec:shape_comparison}),
and study the effect of \RMCC on prediction reliability (\Cref{sec:RMCC_ablation}) and
(iii)~evaluate the hybrid \LiDAR-camera configuration in outdoor environments,
assessing improvements in both scene-level mesh continuity and object-level geometric reconstruction (\Cref{sec:outdoor_hybrid}).

\begin{figure*}[!t]
    \centering
    \begin{subfigure}{0.40\columnwidth}
      \includegraphics[trim={0 0 0 0},clip, width=\columnwidth]{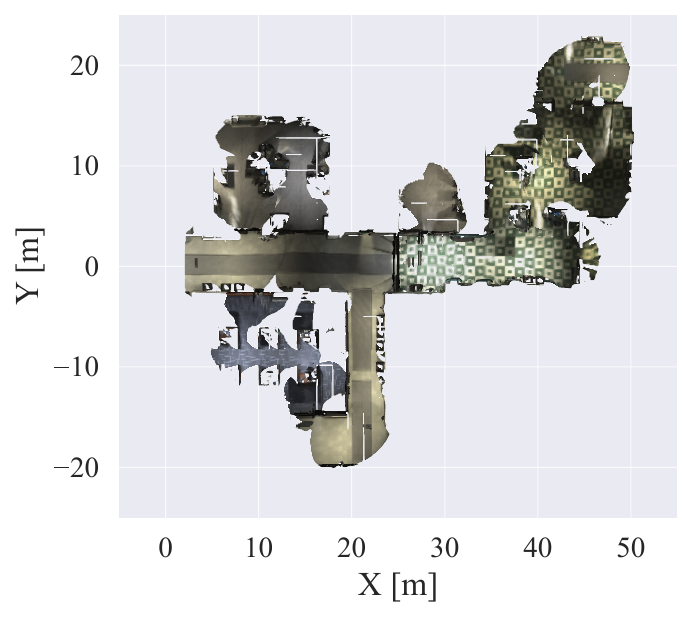}
      \vspace{-1.5em}
      \caption{Reference map}
    \end{subfigure}
    \begin{subfigure}{0.40\columnwidth}
      \includegraphics[trim={0 0 0 0},clip, width=\columnwidth]{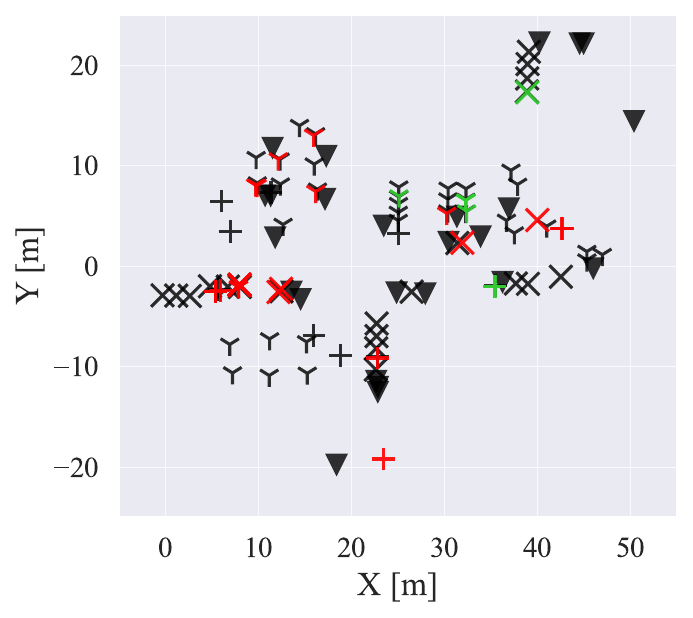}
      \vspace{-1.5em}
      \caption{SlideSLAM~\cite{Liu25tro-SlideSLAM}}
    \end{subfigure}
  	\begin{subfigure}{0.40\columnwidth}
      \includegraphics[trim={0 0 0 0},clip, width=\columnwidth]{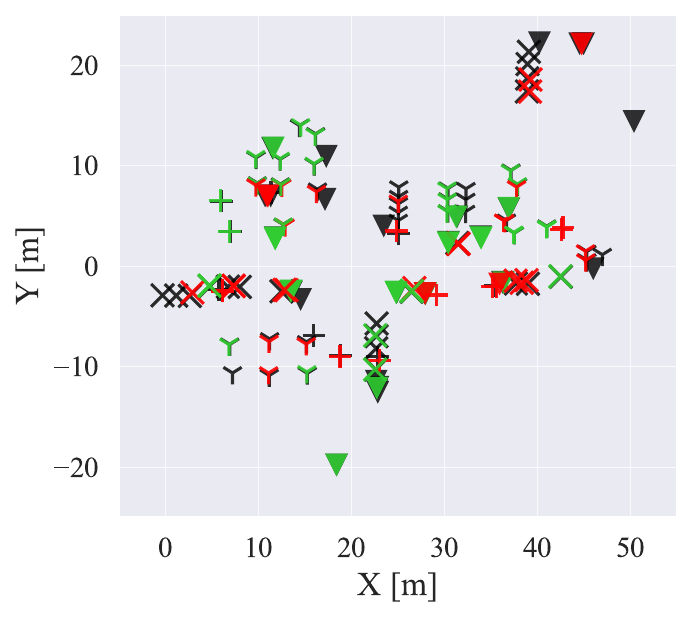}
      \vspace{-1.5em}
      \caption{Hydra~\cite{Hughes24ijrr-hydraFoundations}}
    \end{subfigure}
  	\begin{subfigure}{0.40\columnwidth}
      \includegraphics[trim={0 0 0 0},clip, width=\columnwidth]{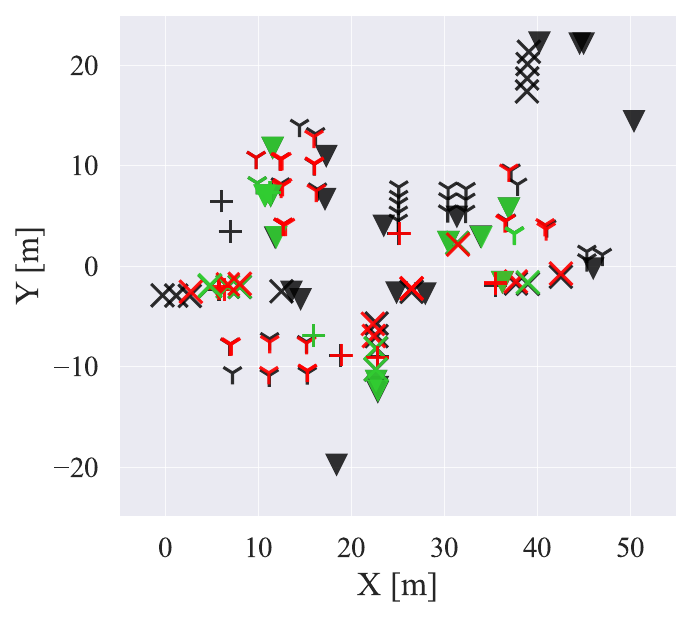}
      \vspace{-1.5em}
      \caption{Khronos~\cite{Schmid24rss-khronos}}
    \end{subfigure}
  	\begin{subfigure}{0.40\columnwidth}
      \includegraphics[trim={0 0 0 0},clip, width=\columnwidth]{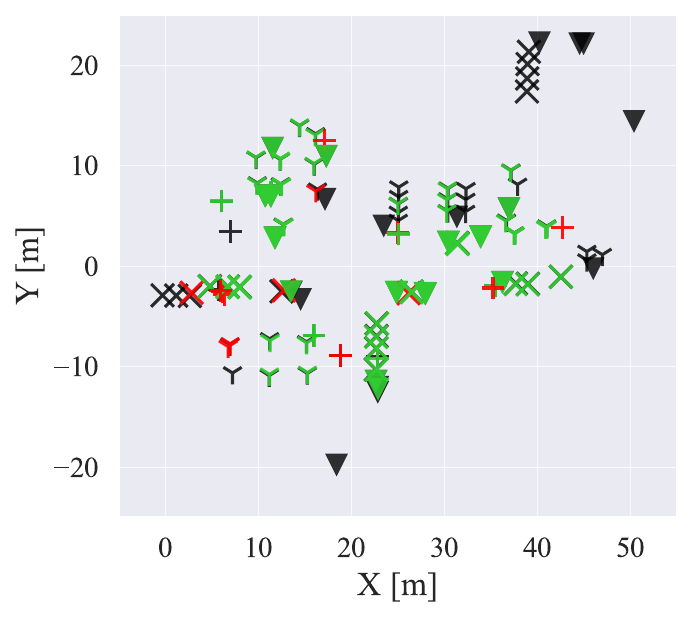}
      \vspace{-1.5em}
      \caption{Hydra++ (Ours)}
    \end{subfigure}
      \includegraphics[trim={0 0 0 0},clip, width=1.75\columnwidth]{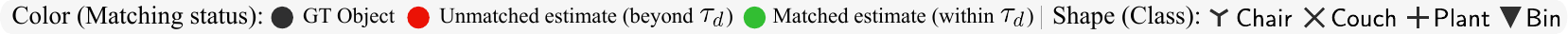}
    \caption{(a) Reference point cloud map and qualitative comparison of object-level mapping results for (b) SlideSLAM~\cite{Liu25tro-SlideSLAM}, (c) Hydra~\cite{Hughes24ijrr-hydraFoundations}, (d) Khronos~\cite{Schmid24rss-khronos}, and (e) our \methodname.
    An estimated object is marked as matched~(green) if the closest ground-truth~(GT) object of the same class lies within $\tau_d=0.2$\,m, and unmatched~(red) otherwise, \ie the centroid of the partially reconstructed object deviates substantially from the GT centroid.}
    \label{fig:map}
    \vspace{-3mm}
\end{figure*}

\subsection{Environment Setups}\label{sec:exp_setup}

First, to compare our approach with state-of-the-art scene graph approaches~\cite{Hughes24ijrr-hydraFoundations,Schmid24rss-khronos},
we utilize the uHumans2 dataset~\cite{Rosinol21ijrr-Kimera}, a photorealistic simulation environment that provides ground-truth object meshes with semantic annotations.
We select \classtype{Chair}, \classtype{Couch}, \classtype{Plant}, and \classtype{Bin} as object categories of interest.

Importantly, in closed-set semantic segmentation, a finite label taxonomy forces many distinct real-world object instances to be mapped to the same category label,
inevitably introducing substantial intra-class variation~\cite{Zhou17cvpr-ade20k}.
For example, in the office scene in uHumans2, the \chair class includes (i) office chairs with armrests and wheels,
(ii) office chairs with wheels, and (iii) table chairs,
while the \bin class includes (i) small bins, (ii) large bins, and (iii) water dispensers.
% This setup intentionally reflects a common deployment regime, 
% where category names capture only a rough semantic grouping rather than a precise shape template.
Consequently, this office scene is well-suited to highlight the benefits of our category-agnostic shape prediction,
which can accommodate large intra-class diversity without relying on category-specific shape priors.

For real-world deployment, we used a sequence from the Kimera-Multi dataset~\cite{Tian23iros-KimeraMultiExperiments}.
The dataset features a robot equipped with a RealSense D455 \RGBD camera and a Velodyne 3D \LiDAR, enabling hybrid sensor configurations.
Training \CRISP on this real-world data requires two inputs for each instance: (a)~a segmentation mask and (b)~a ground-truth object mesh.
For masks, we applied SAM~\cite{Kirillov23iccv-SAM} to automatically segment object instances from the RGB frames,
with the goal of assessing whether \CRISP can be trained using only off-the-shelf tools without manual annotation,
thereby probing its feasibility for in-the-wild deployment.
In total, approximately 50 keyframes were used across three object categories: \classtype{fire hydrant}, \classtype{car}, and \classtype{bench}.

For ground-truth geometry, we scanned target objects using an iPad Pro with a built-in \LiDAR sensor and a 3D scanning application.
The resulting meshes contained holes from limited scanning coverage and thin structures (\eg narrow object parts),
which we manually repaired using Blender~\cite{Blender18-blender} to produce watertight meshes.
This hole-filling step is crucial: \CRISP relies on a well-defined signed distance field,
and holes introduce sign ambiguities near open boundaries that can corrupt the learned shape codes. 

% \Cref{table:hydra_params} summarizes the parameters used in our indoor and outdoor experiments.

% \begin{table}[t!]
% 	\centering
% 	\caption{Main parameters used by \methodname in indoor and outdoor scenes.}
% 	\label{table:hydra_params}
% 	\setlength{\tabcolsep}{3pt}
% 	{\scriptsize
% 		\begin{tabular}{lccc}
% 			\toprule
% 			Param. & Description & Indoor & Outdoor \\ \midrule
%       $v$                    & Voxel resolution                           & 0.1\,m & 0.2\,m \\
%       $\tau_{\text{dyn}}$    & Dynamic detection threshold                & \multicolumn{2}{c}{1\,m} \\
%       $\delta_{\text{grid}}$ & Clustering tolerance                       & \multicolumn{2}{c}{0.1\,m} \\
%       $\tau_{\IoU}$          & IoU threshold (same-label / cross-label)   & 0.25 / 0.1 & 0.1 / 0.3 \\
%       $\tau_{\text{temp}}$   & Temporal window for inactive tracks        & 10 s & 5 s \\
%       $\tau_{\precisionVar}$ & Threshold for \RMCC                        & \multicolumn{2}{c}{0.21} \\
%       \bottomrule
% 		\end{tabular}
% 	}
% \end{table}

%%%%%%%%%%%%%%%%%%%%%%%%%%%%%%%%%%%%%%%%%%%%%%
\subsection{Evaluation Metrics}\label{sec:metrics}

We evaluate object-level mapping in two stages: (i) detection performance and (ii) geometric reconstruction quality.
For detection, we use centroid-based metrics.
Given a distance threshold $\tau_d$,
an estimated object is counted as a true positive if there exists a ground-truth object of the same semantic class whose centroid lies within $\tau_d$ of the estimated centroid.
Thus, unmatched estimates are false positives, and unmatched ground-truth objects are false negatives.
We report precision~(P), recall~(R), and~F$_1$ score.

For geometric reconstruction, we evaluate each matched object pair~(centroid distance $<\tau_d$) using the following metrics:
\begin{itemize}
\item \emph{Chamfer Distance}: Bidirectional point-to-point distance between point sets $\pointSetPred$ and $\pointSetGT$, sampled from predicted and ground-truth meshes, respectively.
% \begin{equation}
% \begin{aligned}
% d_C(\pointSetPred, \pointSetGT)= &\ \frac{1}{|\pointSetPred|} \sum_{\pointPred \in \pointSetPred} \min_{\pointGT \in \pointSetGT}\|\pointPred-\pointGT\|^2 \\
% &\ + \frac{1}{|\pointSetGT|} \sum_{\pointGT \in \pointSetGT} \min_{\pointPred \in \pointSetPred}\|\pointGT-\pointPred\|^2,
% \end{aligned}
% \end{equation}
\item \emph{Bounding-Box IoU (B-IoU)}: Intersection-over-union of the 3D axis-aligned bounding boxes of the matched objects:
\begin{equation}
  \scalebox{0.8}{$\operatorname{B\text{-}IoU}(\cluster_{\subPred}, \cluster_{\subGT})=\dfrac{\Volume\!\left(\BBox(\cluster_{\subPred}) \cap \BBox(\cluster_{\subGT})\right)}{\Volume\!\left(\BBox(\cluster_{\subPred}) \cup \BBox(\cluster_{\subGT})\right)},$}
\label{eq:biou}
\end{equation}
where $\cluster_{\subPred}$ and $\cluster_{\subGT}$ denote the predicted and ground-truth object clusters, respectively.
\item \emph{Volumetric IoU (V-IoU)}: Intersection-over-union between voxelized occupancies of the matched meshes, following the same form as \Cref{eq:biou} with $\voxelSet_{\subPred}$ and $\voxelSet_{\subGT}$ denoting the sets of occupied voxels for the predicted and ground-truth meshes, respectively.
\end{itemize}

\definecolor{myemerald}{rgb}{0.753, 0.898, 0.804}
\definecolor{mylightgreen}{rgb}{0.894, 0.933, 0.745}
\definecolor{myyellow}{rgb}{0.996, 0.972, 0.780}
\newcommand{\cg}{\cellcolor{gray!15}} % Cell color gray
\newcommand{\firstc}{\cellcolor{myemerald!100}}
\newcommand{\secondc}{\cellcolor{mylightgreen!100}}
\newcommand{\thirdc}{\cellcolor{myyellow!100}}

\begin{table}[t!]
    \centering  
    \setlength{\tabcolsep}{4pt}
    \captionsetup{font=footnotesize}
    \caption{Performance comparison of state-of-the-art approaches for a given distance threshold $\tau_d$ in the uHumans2 dataset~\cite{Rosinol21ijrr-Kimera}.
    Precision~(P), recall~(R), and F$_1$ are computed using the centroids of the ground-truth and estimated object bounding boxes.
  Note that Chamfer distance (Chamf.), B-IoU, and V-IoU are averaged only over valid object pairs, 
  where the distance between the estimated and ground-truth centroids is below $\tau_d$.}
    \vspace{-1mm}
    {\scriptsize
    \begin{tabular}{l|l|ccc|ccc}
        \toprule \midrule
        % This case, voxel size of v-IoU is 0.0375
        % Method  &  {P\,$\uparrow$}  &   {R\,$\uparrow$} & F$_1$\,$\uparrow$ & Chamf. & B-IoU  & V-IoU \\ \midrule
        % Hydra         & 0.529 & 0.409 & 0.461 & 0.020 & 0.514 & 0.250 \\ 
        % Khronos        & 0.362 & 0.239 & 0.288 & 0.004 & 0.671 & 0.318  \\ 
        % Ours w/o \RMCC   & 0.670 & 0.602 & 0.634 & 0.003 & 0.690 & 0.458 \\ 
        % Ours             & 0.843 & 0.591 & 0.695 & 0.003 & 0.702 & 0.467 \\ \midrule \bottomrule 
       & Method  &  {P\,$\uparrow$}  &   {R\,$\uparrow$} & F$_1$\,$\uparrow$ & Chamf. $\downarrow$  & B-IoU $\uparrow$  & V-IoU $\uparrow$ \\ \midrule
       \parbox[t]{2mm}{\multirow{5}{*}{\rotatebox[origin=c]{90}{$\tau_d = 0.2$\,m}}}
       & SlideSLAM            & 0.206 & 0.057 & 0.089 & 0.039 & 0.606 & 0.199 \\ 
       % & SlideSLAM$\dagger$   & 0.206 & 0.079 & 0.115 & 0.039 & 0.605 & 0.201 \\ 
       & Hydra          & \thirdc 0.529 & \thirdc 0.409 & \thirdc 0.461 & 0.035 & 0.515 & 0.298 \\ 
       & Khronos        & 0.362 & 0.239 & 0.288 & \thirdc 0.028 & \thirdc 0.671 & \thirdc 0.350  \\ 
       & Ours w/o \RMCC & \secondc 0.670 & \firstc \textbf{0.602} & \secondc 0.634 & \secondc 0.008 & \secondc 0.690 & \secondc 0.540 \\ 
       & Ours           & \firstc \textbf{0.787} & \secondc 0.591 & \firstc \textbf{0.675} & \firstc \textbf{0.004} & \firstc \textbf{0.739} & \firstc \textbf{0.598} \\ \midrule
       \parbox[t]{2mm}{\multirow{5}{*}{\rotatebox[origin=c]{90}{$\tau_d = 0.5$\,m}}}
       & SlideSLAM      & 0.794 & 0.159 & 0.265 & \thirdc 0.051 & \thirdc 0.436 & 0.134 \\
       % & SlideSLAM$\dagger$ & 0.7941 & 0.2222 & 0.3473 & 0.051 & 0.436 & 0.134 \\
       & Hydra          & \thirdc 0.897 & \firstc \textbf{0.705} & \firstc \textbf{0.789} & 0.072 & 0.412 & \thirdc 0.246 \\
       & Khronos        & \firstc \textbf{0.971} & 0.546 & 0.699 & 0.091 & 0.358 & 0.239  \\ 
       & Ours w/o \RMCC & 0.839 & \secondc 0.682 & \thirdc 0.752 & \secondc 0.032 & \secondc 0.582 & \secondc 0.451 \\
       & Ours           & \secondc 0.921 & \thirdc 0.671 & \secondc 0.776 & \firstc \textbf{0.020} & \firstc \textbf{0.655} & \firstc \textbf{0.528} \\ \midrule \bottomrule
    \end{tabular}
    }
    \label{table:comparison}
    \vspace{-1mm}
\end{table}

\subsection{Object-Level Scene Graph Quality}\label{sec:sg_quality}

First, we compare \methodname against existing scene graph SLAM approaches: Hydra~\cite{Hughes24ijrr-hydraFoundations},
which clusters semantic meshes to extract objects,
and Khronos~\cite{Schmid24rss-khronos}, which reconstructs objects with instance segmentation, on the uHumans2 dataset.

\begin{figure}[t]
    \centering
    % ===== Column (a): SlideSLAM =====
    \begin{subfigure}{0.11\textwidth}
        % Row 05
        \includegraphics[width=0.495\linewidth]{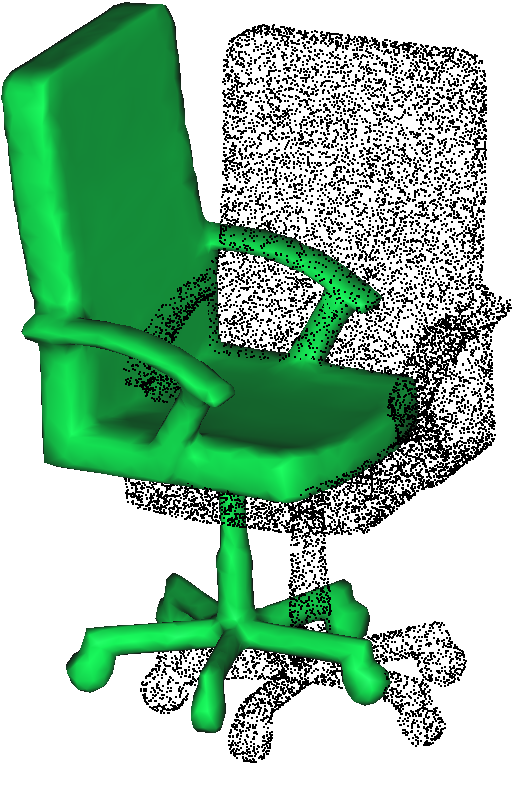}%
        \includegraphics[width=0.495\linewidth]{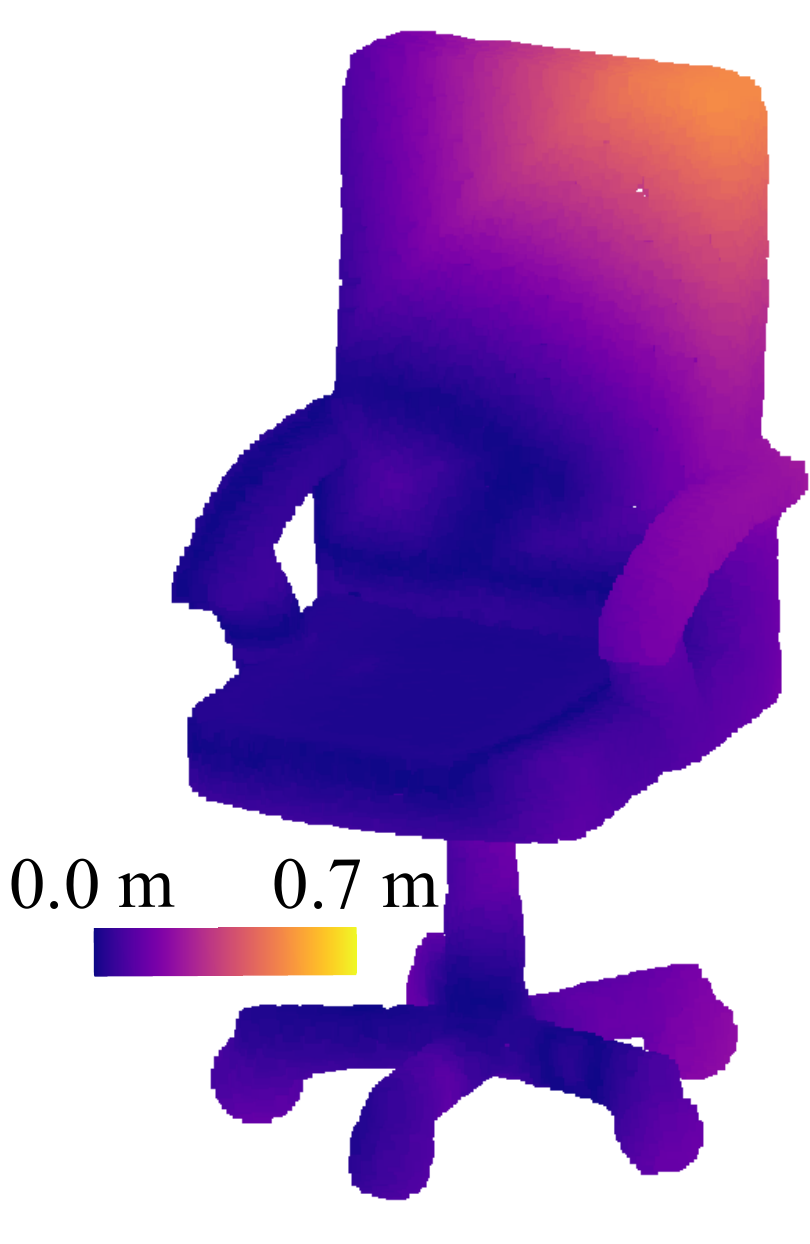}\\[0.0em]
        \includegraphics[width=0.495\linewidth]{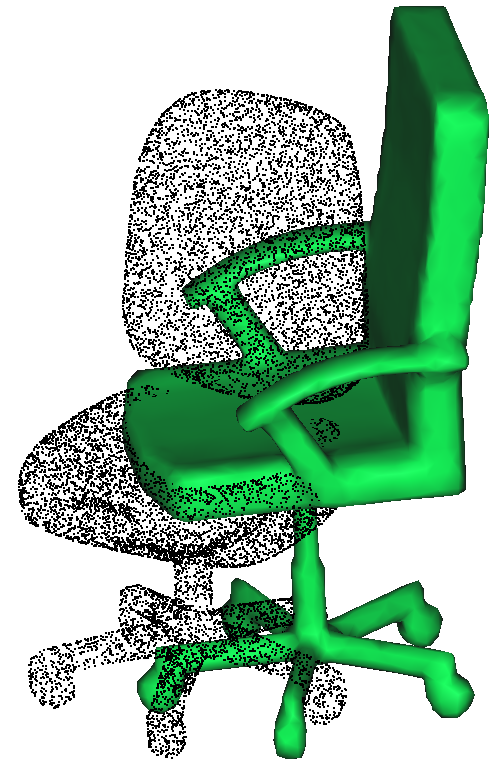}%
        \includegraphics[width=0.495\linewidth]{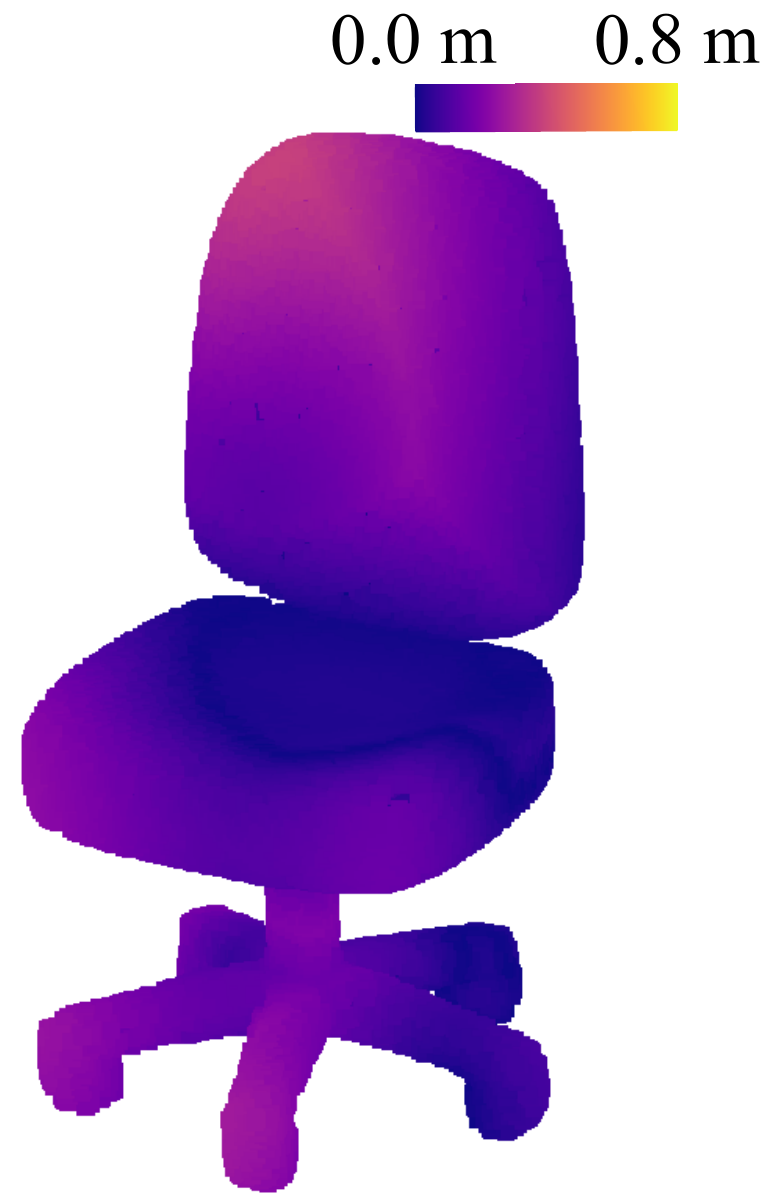}\\[0.0em]
        \includegraphics[width=0.495\linewidth]{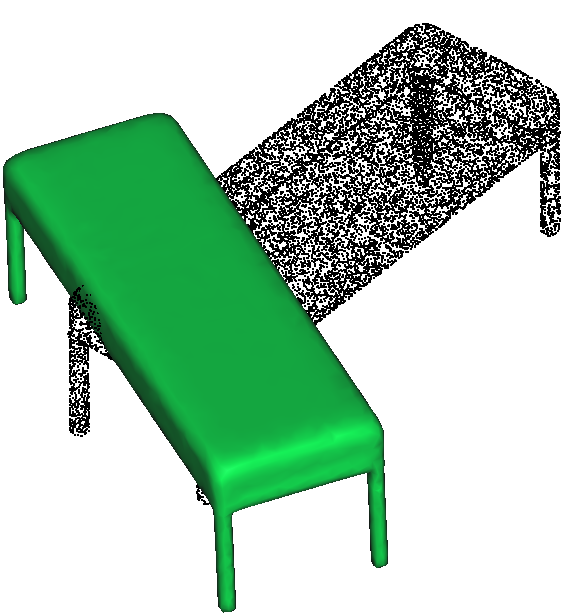}%
        \includegraphics[width=0.495\linewidth]{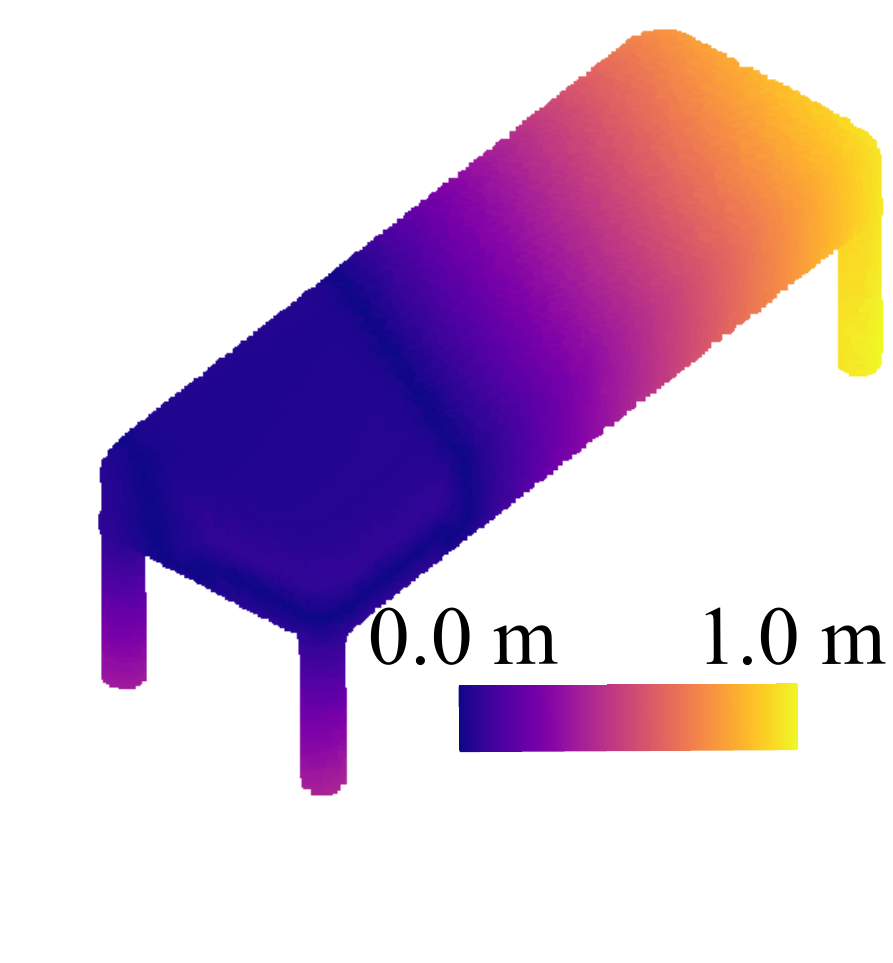}\\[0.0em]
        \includegraphics[width=0.495\linewidth]{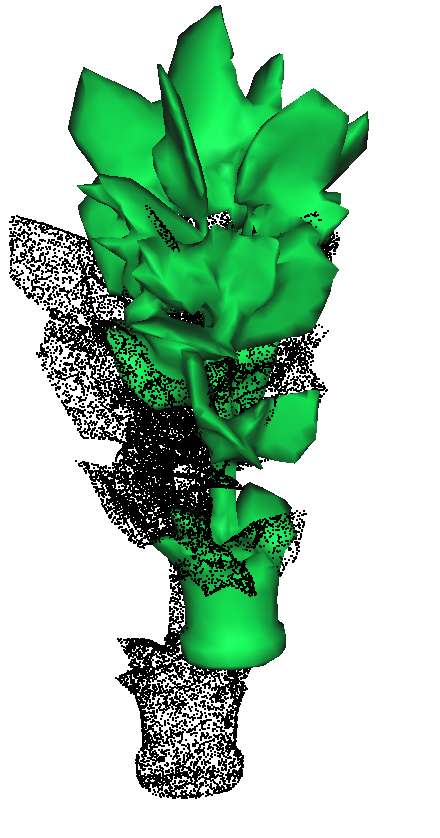}%
        \includegraphics[width=0.495\linewidth]{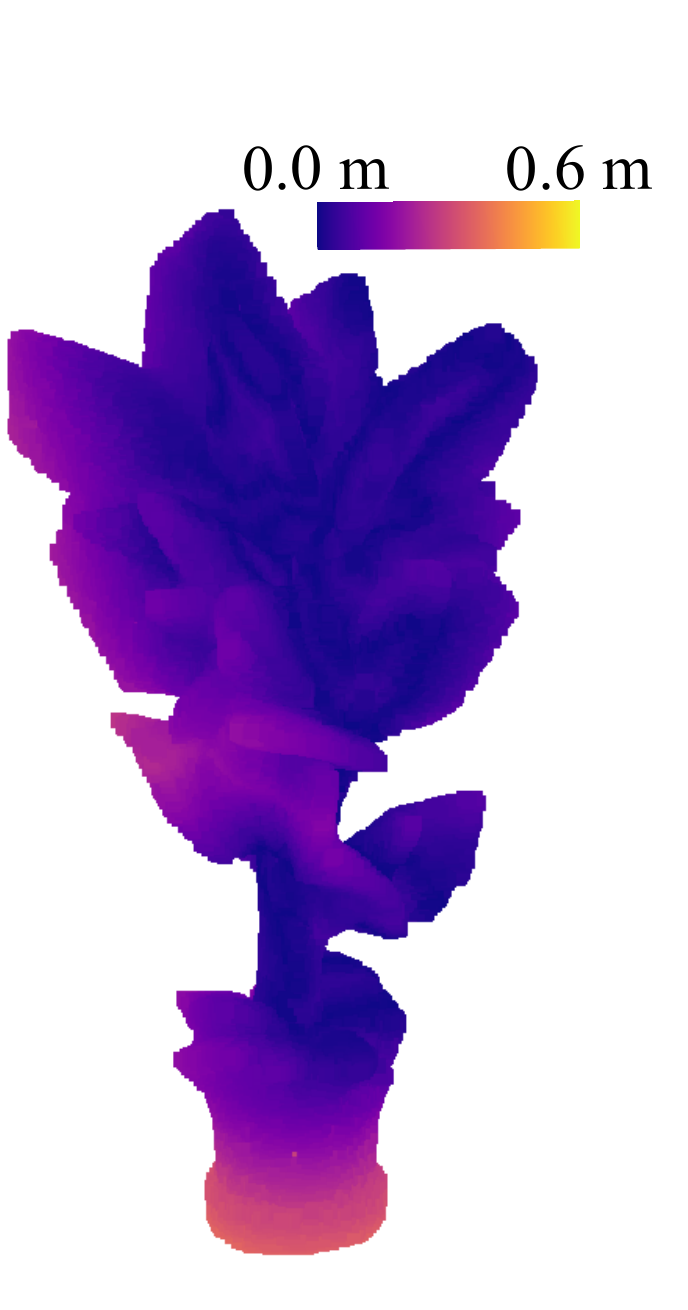}
        \caption{\scalebox{0.94}{SlideSLAM~\cite{Liu25tro-SlideSLAM}}}
    \end{subfigure}
    \hfill
    % ===== Column (b): Mesh Objects =====
    \begin{subfigure}{0.11\textwidth}
        \includegraphics[width=0.495\linewidth]{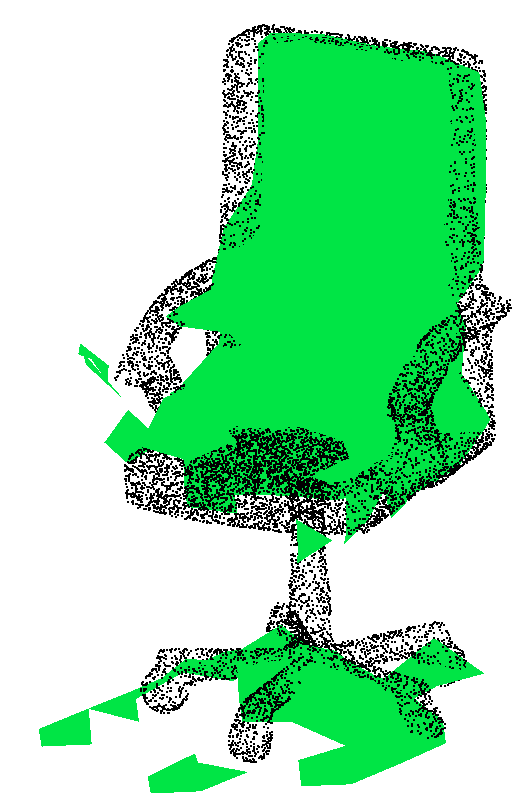}%
        \includegraphics[width=0.495\linewidth]{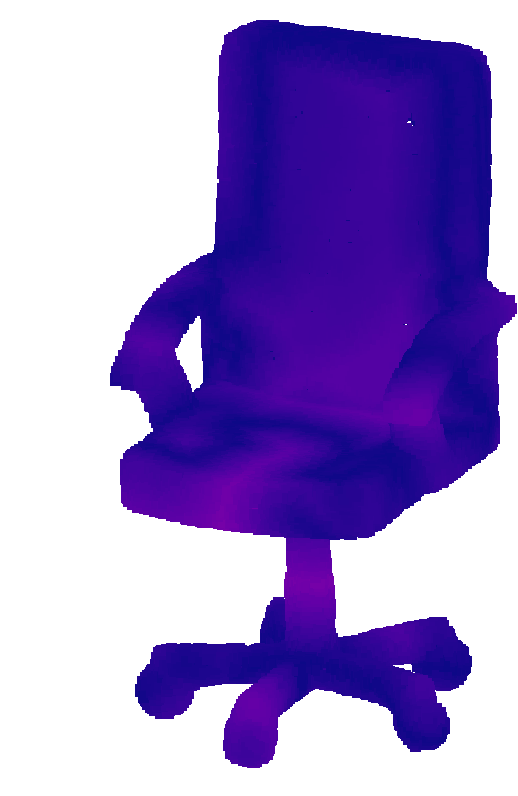}\\[0.0em]
        \includegraphics[width=0.495\linewidth]{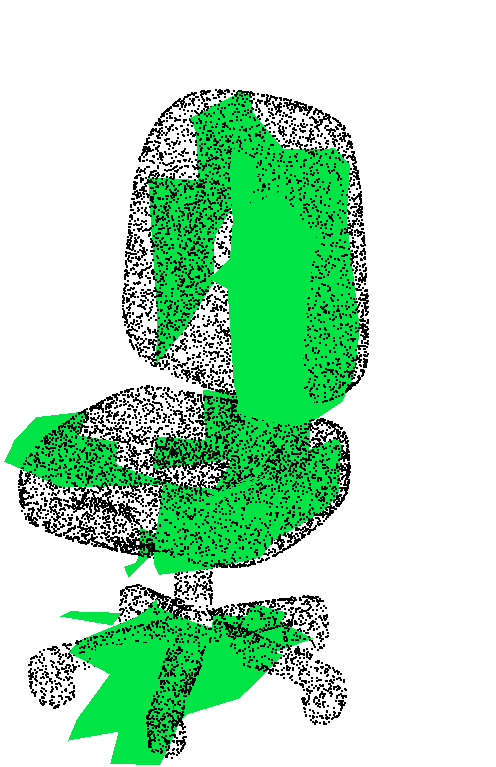}%
        \includegraphics[width=0.495\linewidth]{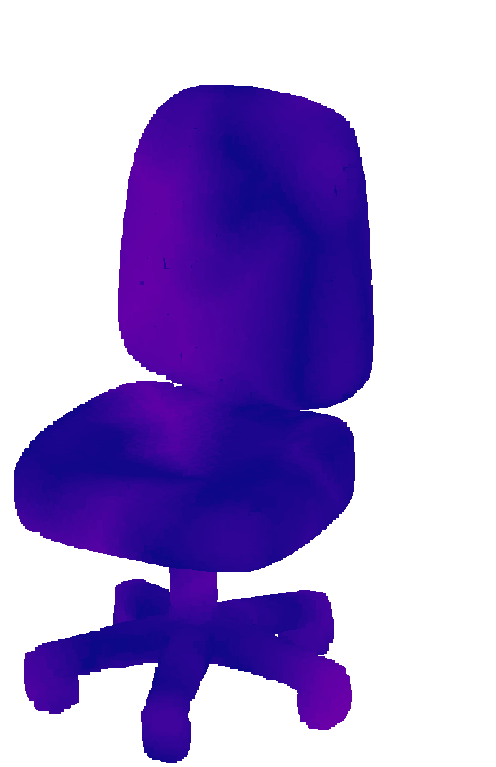}\\[0.0em]
        \includegraphics[width=0.495\linewidth]{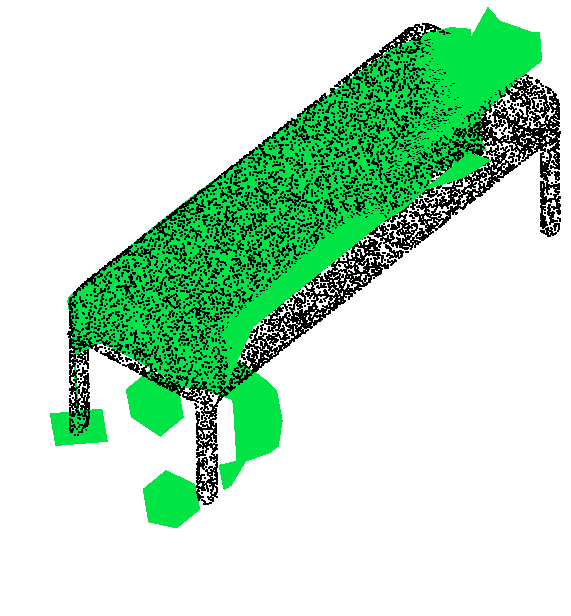}%
        \includegraphics[width=0.495\linewidth]{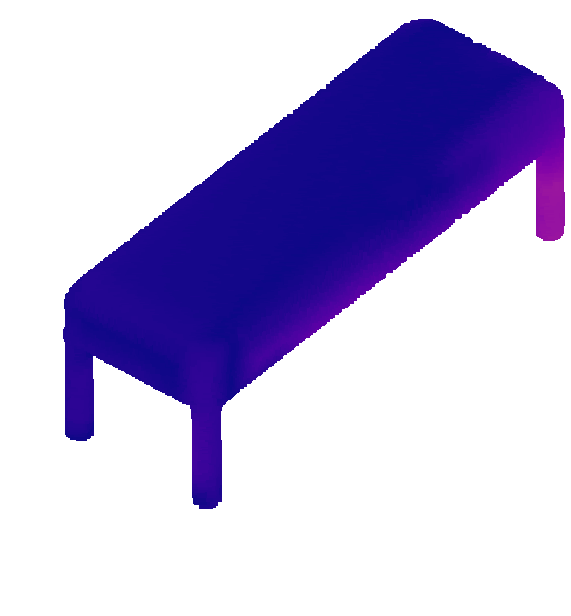}\\[0.0em]
        \includegraphics[width=0.495\linewidth]{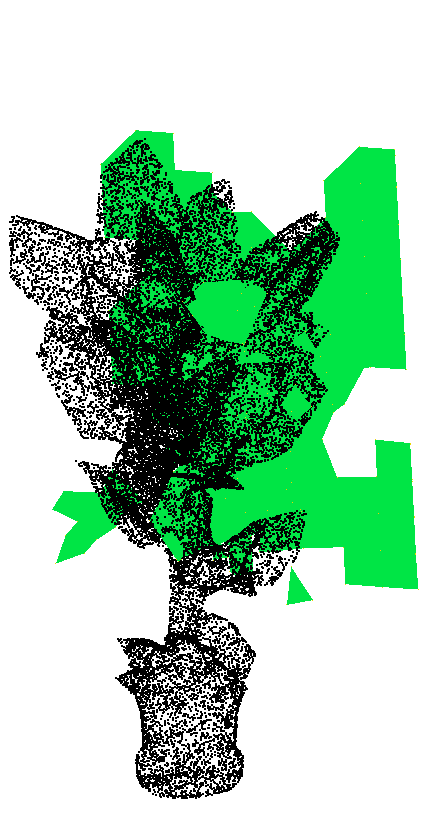}%
        \includegraphics[width=0.495\linewidth]{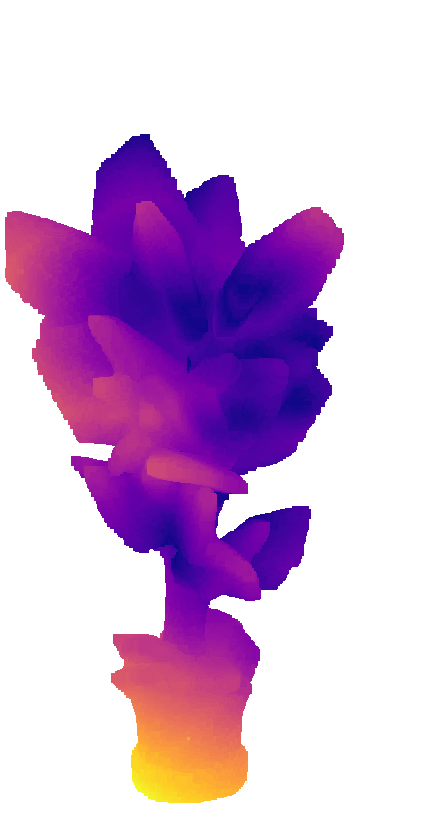}
        \caption{Hydra~\cite{Hughes24ijrr-hydraFoundations}}
    \end{subfigure}
    \hfill
    % ===== Column (c): Khronos =====
    \begin{subfigure}{0.11\textwidth}
        \includegraphics[width=0.495\linewidth]{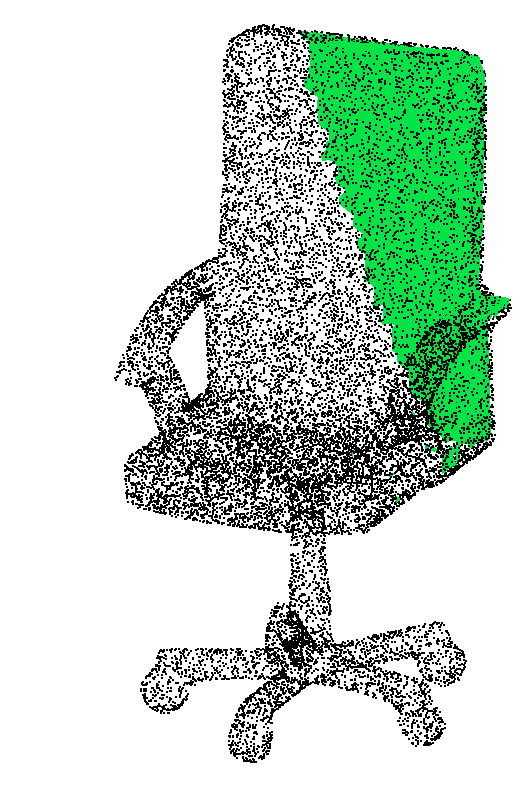}%
        \includegraphics[width=0.495\linewidth]{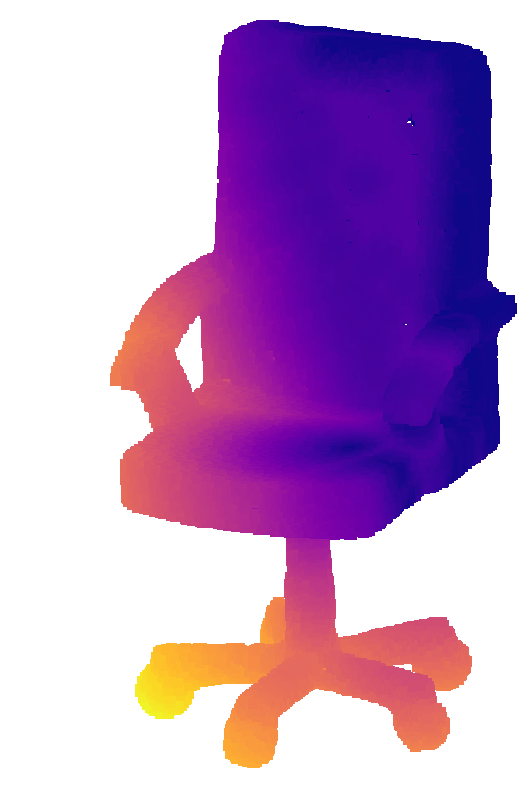}\\[0.0em]
        \includegraphics[width=0.495\linewidth]{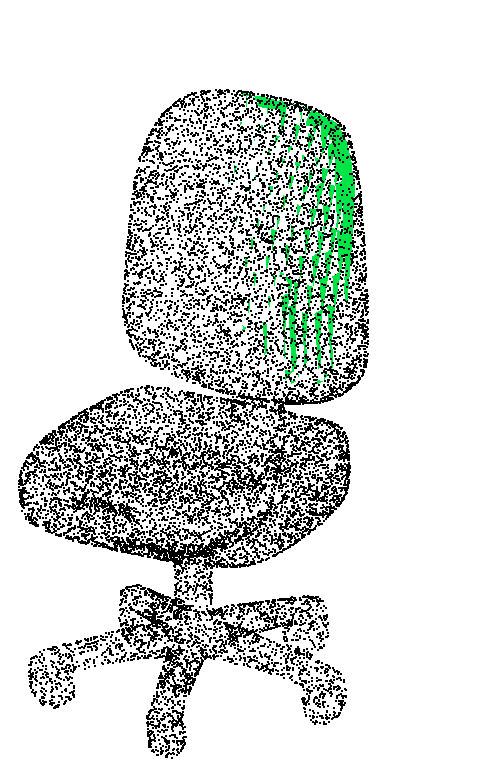}%
        \includegraphics[width=0.495\linewidth]{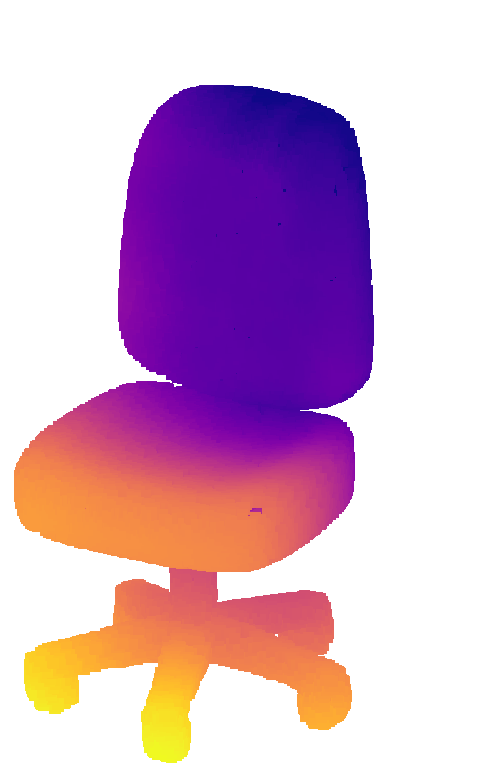}\\[0.0em]
        \includegraphics[width=0.495\linewidth]{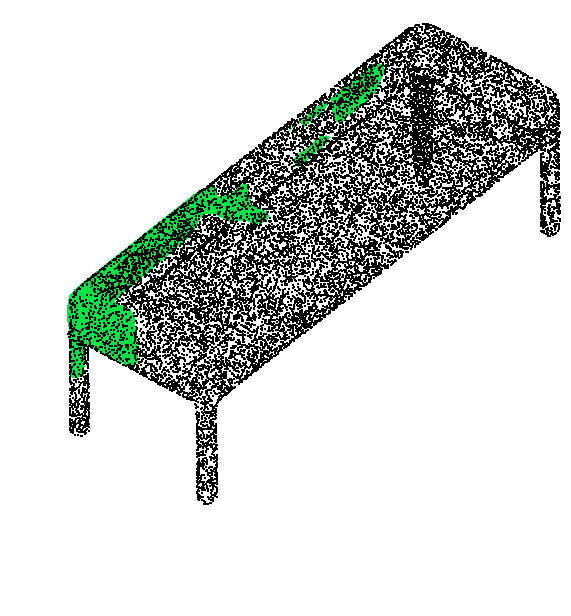}%
        \includegraphics[width=0.495\linewidth]{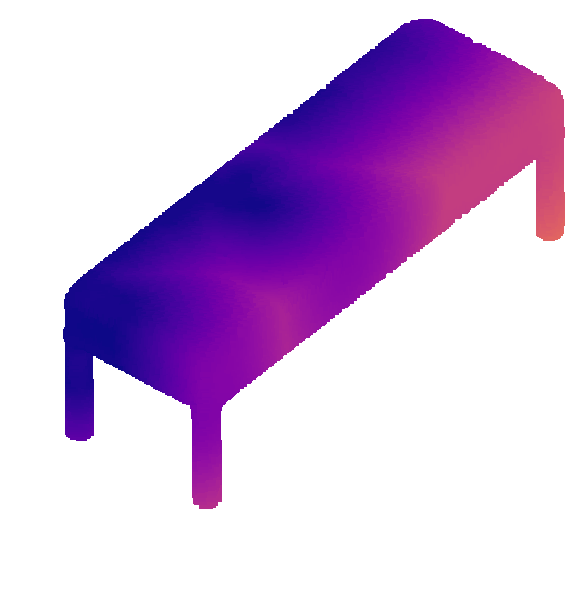}\\[0.0em]
        \includegraphics[width=0.495\linewidth]{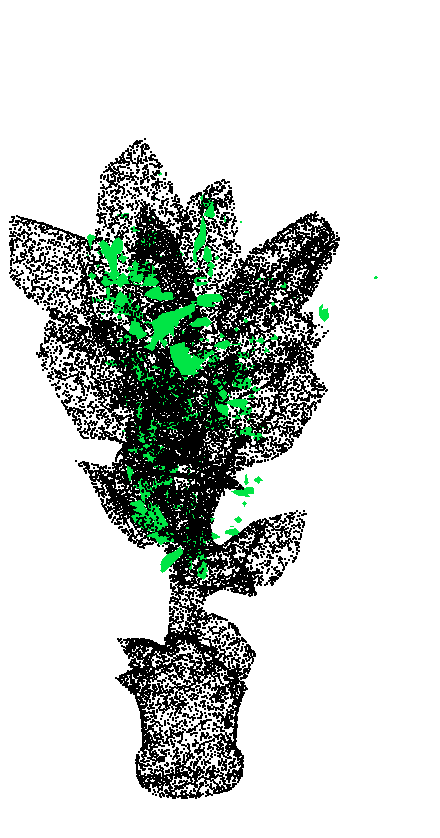}%
        \includegraphics[width=0.495\linewidth]{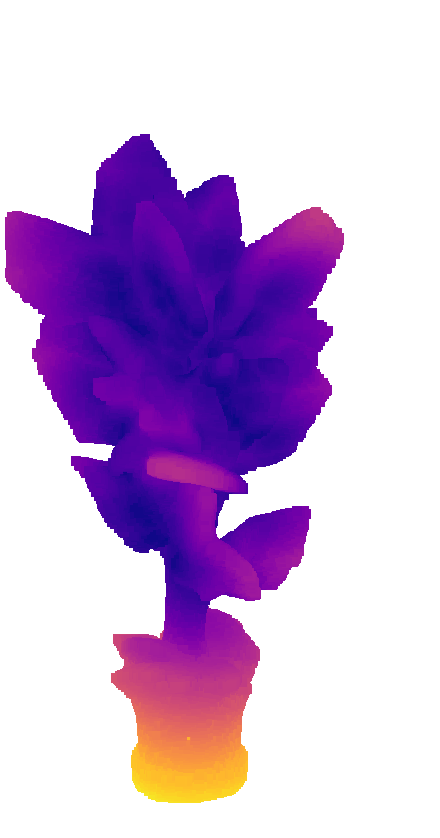}
        \caption{Khronos~\cite{Schmid24rss-khronos}}
    \end{subfigure}
    \hfill
    % ===== Column (d): Crisp =====
    \begin{subfigure}{0.11\textwidth}
        \includegraphics[width=0.495\linewidth]{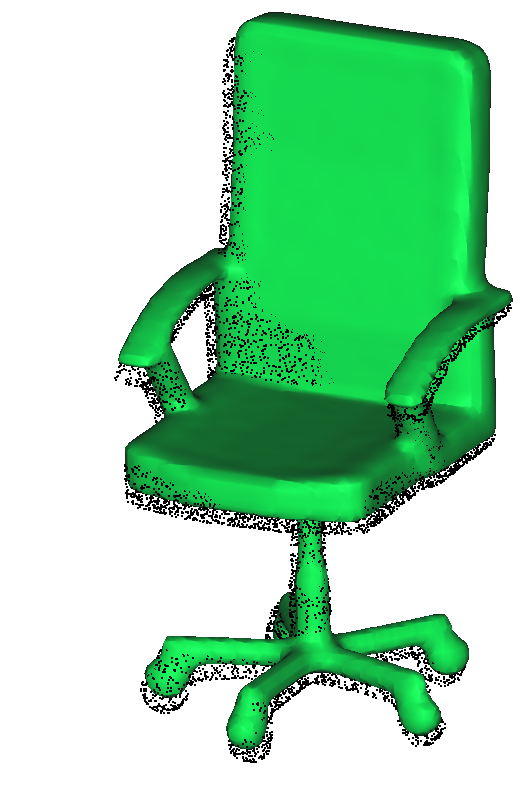}%
        \includegraphics[width=0.495\linewidth]{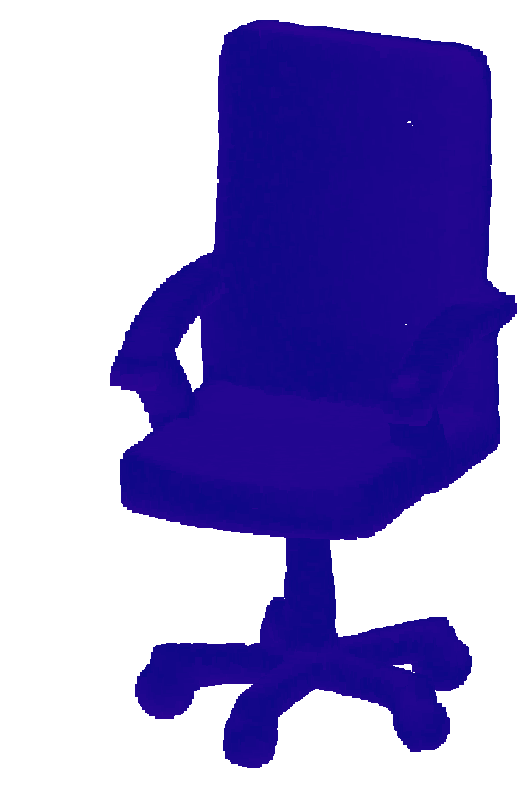}\\[0.0em]
        \includegraphics[width=0.495\linewidth]{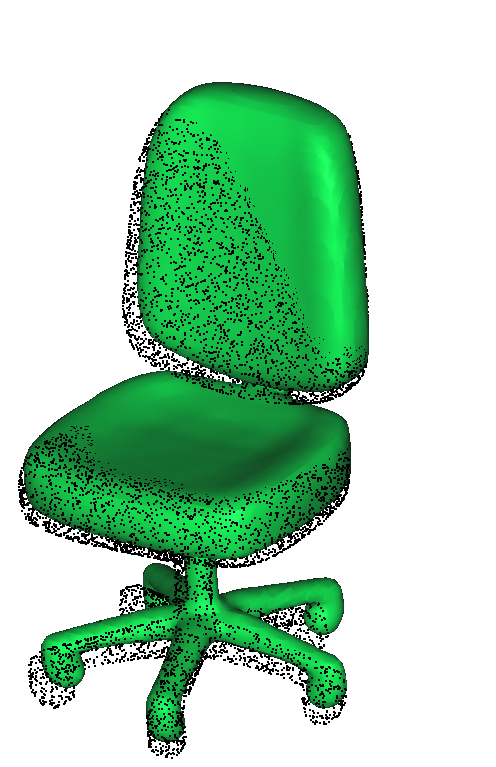}%
        \includegraphics[width=0.495\linewidth]{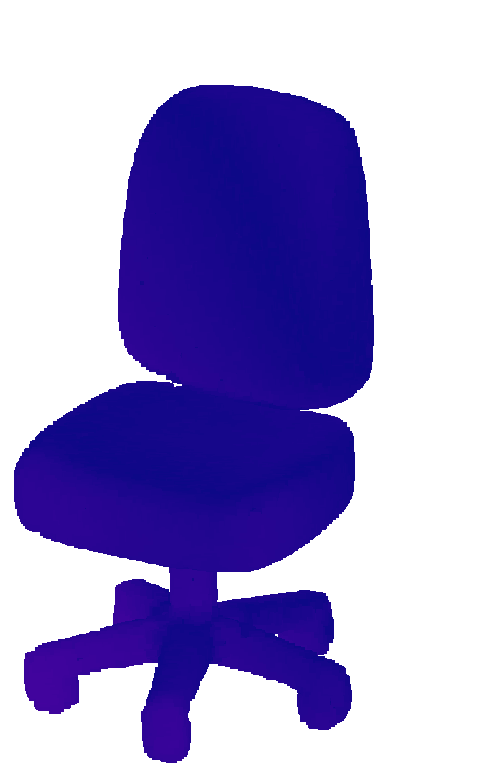}\\[0.0em]
        \includegraphics[width=0.495\linewidth]{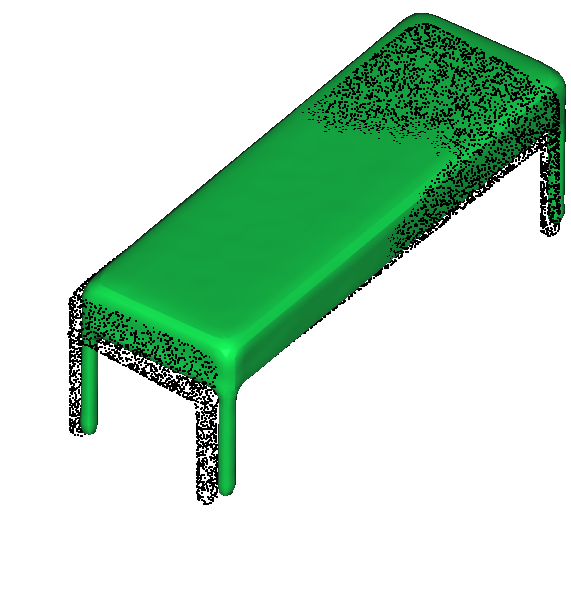}%
        \includegraphics[width=0.495\linewidth]{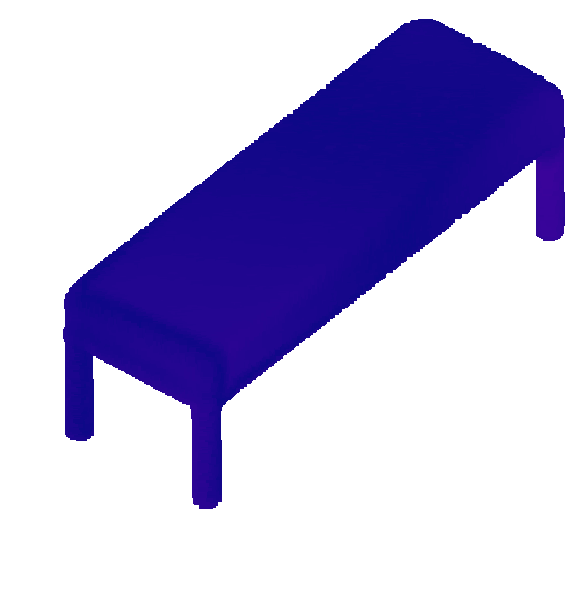}\\[0.0em]
        \includegraphics[width=0.495\linewidth]{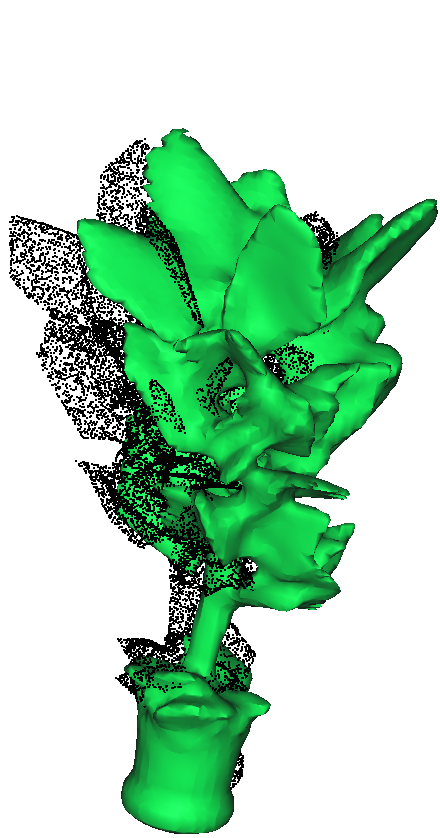}%
        \includegraphics[width=0.495\linewidth]{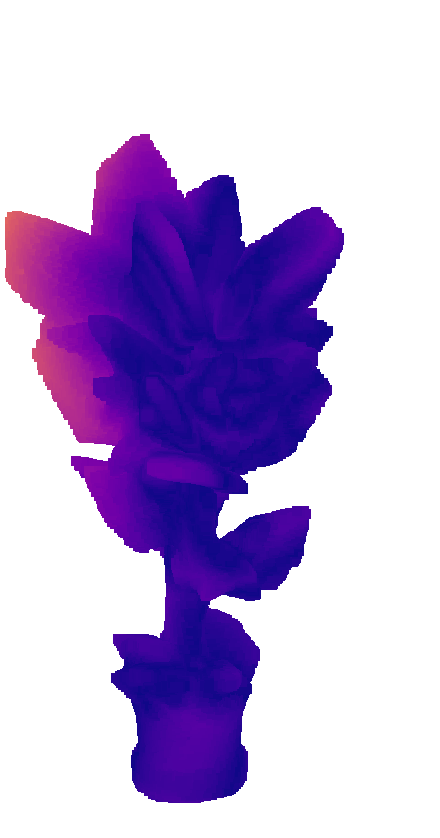}
        \caption{\methodname~(Ours)}
    \end{subfigure}
    \caption{Qualitative comparison with other object representations.
    In each row, the left image shows the estimated mesh~(green) overlaid with ground-truth surface points~(black),
    while the right image shows the corresponding per-point error heatmap.}
    \label{fig:mesh_heatmap_grid}
    % \vspace{-0.1cm}
\end{figure}

As shown in \Cref{fig:map} and \Cref{table:comparison},
\methodname achieves competitive performance across detection and shape metrics at the strict threshold~($\tau_d\!=\!0.2$\,m).
In particular, \Cref{fig:map} shows that although the baselines~\cite{Hughes24ijrr-hydraFoundations,Schmid24rss-khronos} successfully detect objects,
they represent them solely by accumulating partial observations,
which often results in centroid errors exceeding 0.2\,m compared to the ground truth~(the partial meshes can be seen in \Cref{fig:mesh_heatmap_grid}).

At the relaxed threshold~($\tau_d\!=\!0.5$\,m), Hydra achieves the best F$_1$ score by achieving the highest recall,
yet exhibits substantially higher Chamfer distance and lower B-IoU and V-IoU.
This reflects a fundamental limitation of prior approaches: 
because they rely on partial observations for clustering~\cite{Hughes24ijrr-hydraFoundations} and instance tracking~\cite{Schmid24rss-khronos},
they cannot infer geometry in unobserved regions, which ultimately degrades geometric accuracy in the scene graph.
By contrast, \methodname intrinsically performs object shape completion, 
resulting in lower Chamfer distances and higher volumetric overlap.

\subsection{Analysis on Intra-Class Shape Variation}\label{sec:intraclass}

Next, we emphasize the importance of category-agnostic, learning-based shape estimation in handling intra-class variation.
CAD-template-based approaches such as SlideSLAM~\cite{Liu25tro-SlideSLAM} rely on a single canonical shape per semantic class,
which inherently limits their ability to represent diverse real-world instances.
As shown in the first and second rows of \Cref{fig:mesh_heatmap_grid},
visually distinct objects, such as office chairs with and without armrests, are forced to share the same template,
leading to systematic misalignment and increased Chamfer distance when the true geometry deviates from the canonical model.
In contrast, \methodname leverages a learning-based shape estimator to reconstruct instance-specific geometry, capturing salient details such as armrests,
seat curvature, and backrest shape.

Overall, these results demonstrate that fixed per-class CAD templates are insufficient in diverse environments,
whereas \methodname effectively captures intra-class variation through instance-level shape estimation.

% \begin{figure}[!t]
%     \centering
%     \includegraphics[trim={0 0 0 0},clip, width=0.45\columnwidth]{figures/tbu.pdf}
%     \includegraphics[trim={0 0 0 0},clip, width=0.45\columnwidth]{figures/tbu.pdf}
%     \caption{Mapping results over time with (a) SAM3D~\cite{Chen2026CVPR-sam3d} and (b) CRISP~\cite{Shi25cvpr-CRISP}. Note that due to the high inference latency of SAM3D, it is difficult to perform real-time object shape estimation alongside hierarchical scene graph construction, whereas systems using CRISP can operate in near real-time.}
%     \label{fig:mapping_comparison}
% \end{figure}

\begin{figure}[!t]
    \centering
  	\begin{subfigure}{0.9\columnwidth}
      \centering
      \includegraphics[trim={0 0 0 0},clip, width=1.0\columnwidth]{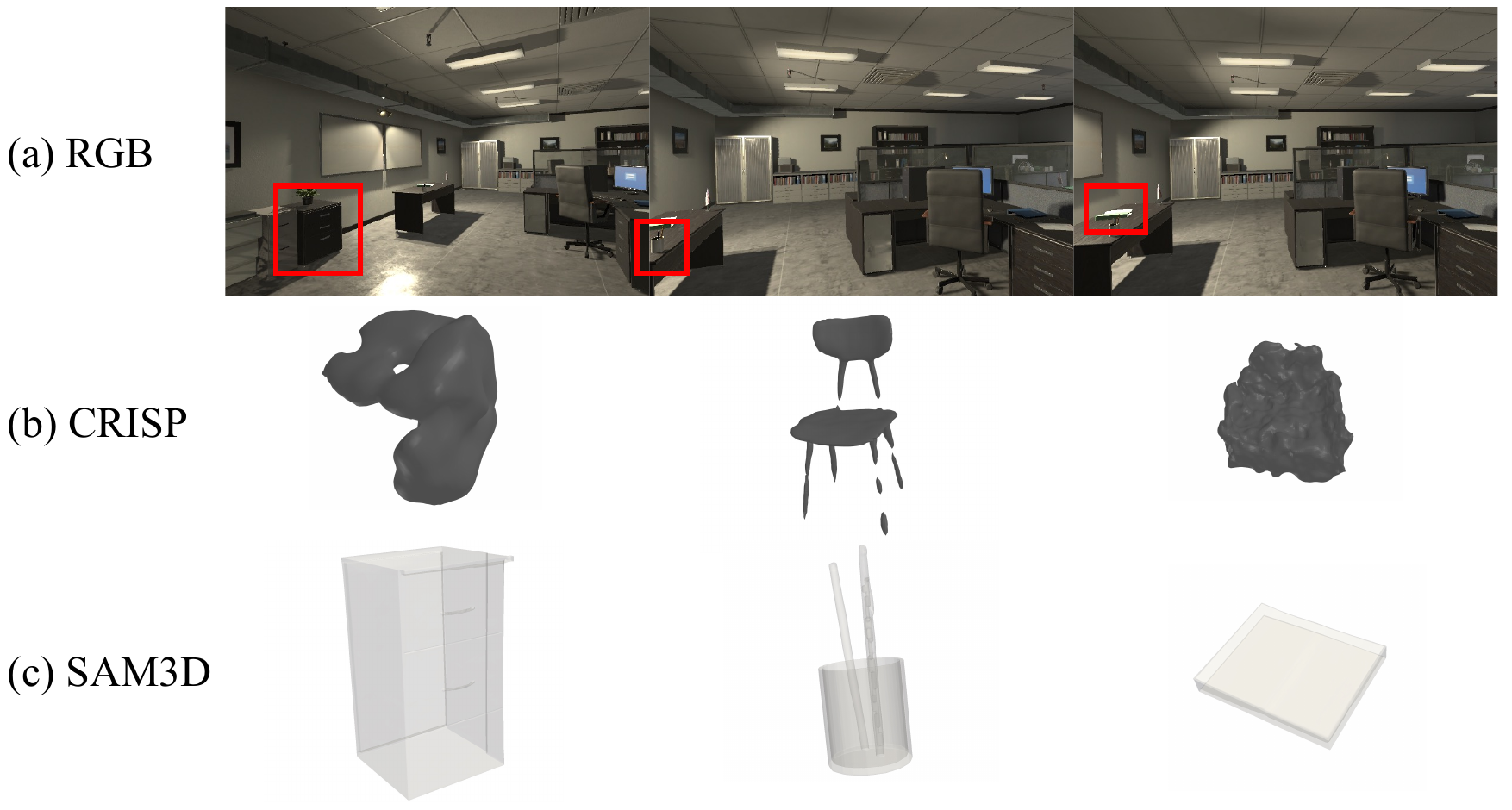}
   	\end{subfigure}
    \vspace{-1mm}
    \caption{Qualitative comparison of mesh reconstruction in an out-of-domain setting, where unseen object instances are given as inputs.
    (a) RGB image for reference, with target objects highlighted in red boxes; (b)~mesh results produced by \CRISP~\cite{Shi25cvpr-CRISP} and (c)~those produced by SAM3D~\cite{Chen2026CVPR-sam3d}.}
      \label{fig:crisp_and_sam3d}
\end{figure}

\begin{figure}[!t]
    \centering
    \includegraphics[trim={0 0 0 0},clip, width=0.45\columnwidth]{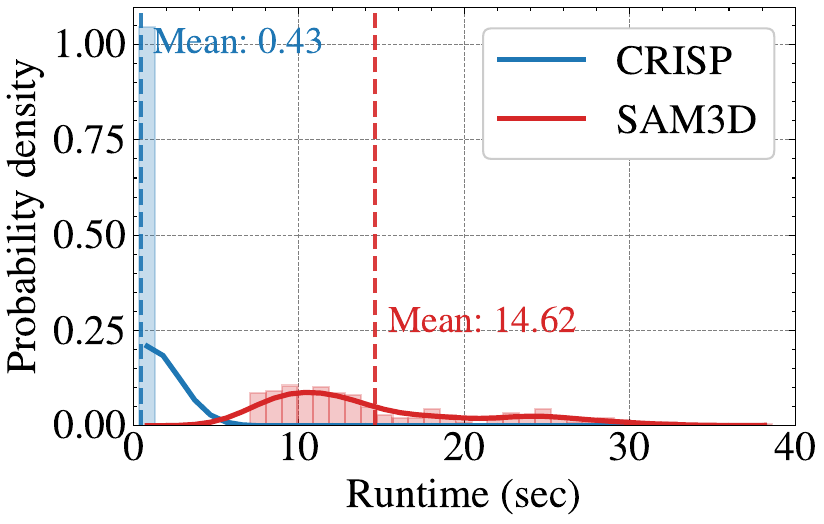}
    \includegraphics[trim={0 0 0 0},clip, width=0.45\columnwidth]{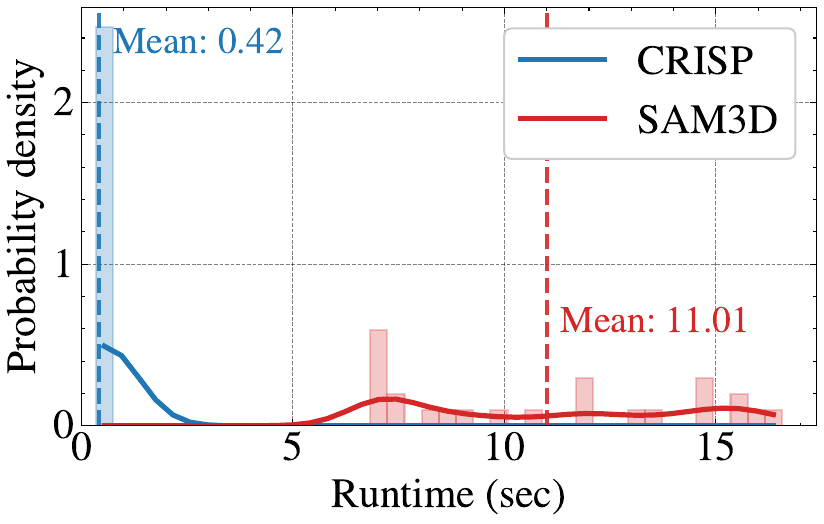}
    \vspace{-1mm}
    \caption{Left to right: Timing comparison between SAM3D~\cite{Chen2026CVPR-sam3d} and CRISP~\cite{Shi25cvpr-CRISP} in indoor (using the uHumans2 dataset~\cite{Rosinol21ijrr-Kimera}) and outdoor (using the Kimera-Multi dataset~\cite{Tian23iros-KimeraMultiExperiments}) environments.}
    \label{fig:timing}
\end{figure}

\subsection{Shape Estimation Model Comparison}\label{sec:shape_comparison}

Although our pipeline is primarily built upon \CRISP~\cite{Shi25cvpr-CRISP},
we further analyze how the choice of shape estimation model influences overall system behavior by comparing \CRISP and SAM3D~\cite{Chen2026CVPR-sam3d}.
This comparison does not imply that every supported estimator enables real-time operation:
\CRISP is the default estimator used for the real-time configuration, whereas SAM3D is evaluated to demonstrate the modularity of the interface and the computational cost associated with stronger out-of-domain generalization.
The key distinction lies in the trade-off between out-of-domain generalization and inference speed.
To test out-of-domain generalization, we introduce an additional
\classtype{computer} class alongside our original four categories.
In the uHumans2 dataset, various distinct office supplies are
ambiguously grouped into this single \classtype{computer} label.

As a foundation model, SAM3D exhibits strong generalization capabilities, reliably reconstructing novel,
out-of-domain geometries where \CRISP struggles (see Figs.~\ref{fig:crisp_and_sam3d}(b) and \ref{fig:crisp_and_sam3d}(c)).
However, this robustness comes at the expense of significantly higher inference latency, as presented in \Cref{fig:timing}; in our experiments, SAM3D requires roughly 11--14\,s per object and is therefore unsuitable for real-time multi-object deployment in the current system.
Conversely, while \CRISP is not a foundation model and is limited to its training distribution, it boasts much faster inference times.

In summary, \CRISP is optimized for real-time deployments over known object classes, 
whereas SAM3D can improve reconstruction of novel objects when the application can tolerate delayed object-level mesh updates.

% In summary, the two models present a practical trade-off: \CRISP is more suitable for deployments requiring real-time operation over known object classes,
% while SAM3D offers better generalization to novel instances at the cost of higher latency.

\subsection{Outdoor Deployment With Hybrid Sensor Fusion Mode}\label{sec:outdoor_hybrid}

Lastly, we demonstrate that our hybrid \LiDAR-camera mode improves both scene-level geometry and object-level reconstruction quality.
To validate this capability beyond indoor \RGBD settings, we deploy \methodname in outdoor environments using a hybrid sensor configuration,
where a 3D \LiDAR point cloud provides metric scene structure while an RGB/\RGBD camera enables \CRISP-based object shape estimation.

\begin{figure}[!t]
    \centering
  	\begin{subfigure}{0.30\columnwidth}
      \includegraphics[trim={0 0 0 0},clip, width=\columnwidth]{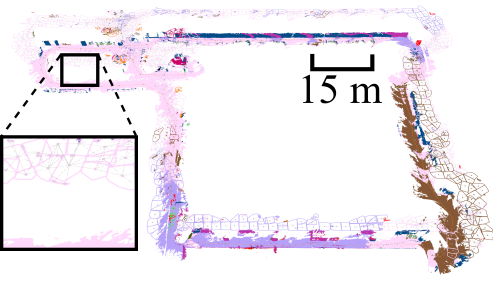}
      \caption{}
   	\end{subfigure}
  	\begin{subfigure}{0.30\columnwidth}
      \includegraphics[trim={0 0 0 0},clip, width=\columnwidth]{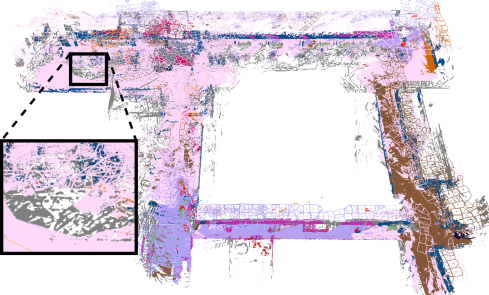}
      \caption{}
   	\end{subfigure}
  	\begin{subfigure}{0.30\columnwidth}
    \includegraphics[trim={0 0 0 0},clip, width=\columnwidth]{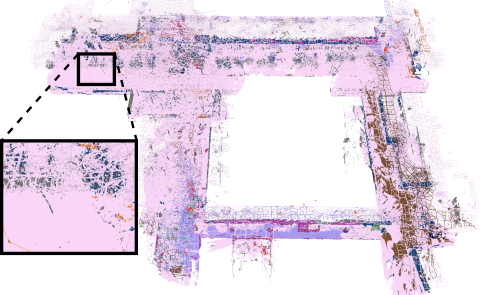}
        \caption{}
   	\end{subfigure}
    \caption{Mesh quality comparison in outdoor scenes: 
      (a) Meshes generated in RGB-D-only mode, corresponding to the original metric-semantic mesh generation in Hydra~\cite{Hughes24ijrr-hydraFoundations} or Khronos~\cite{Schmid24rss-khronos}.
    (b)--(c) Meshes from a hybrid \LiDAR-camera mode without and with the adaptive integration strategy, respectively.}
    \vspace{-3mm}
    \label{fig:mesh_comparison}
\end{figure}

\begin{figure}[!t]
    \centering
  	\begin{subfigure}{0.27\columnwidth}
      \includegraphics[trim={0 0 0 0},clip, width=\columnwidth]{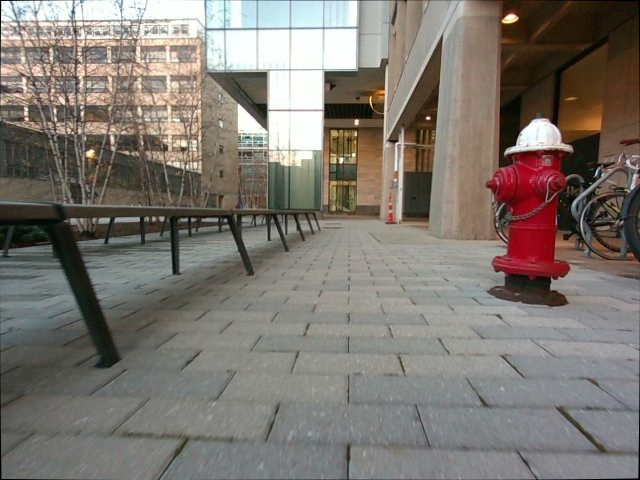}
      \includegraphics[trim={0 0 0 0},clip, width=\columnwidth]{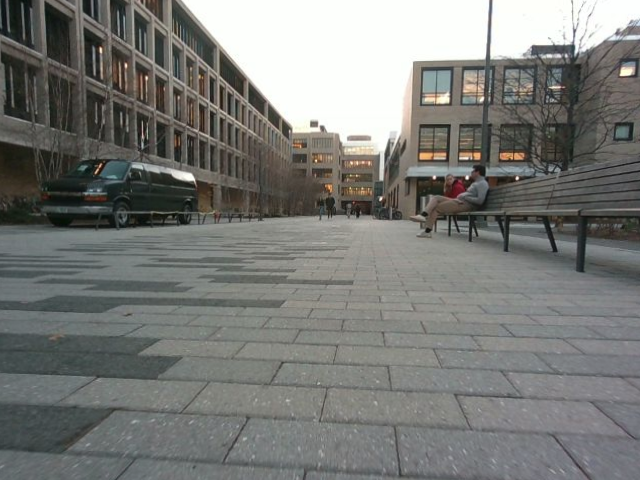}
      \includegraphics[trim={0 0 0 0},clip, width=\columnwidth]{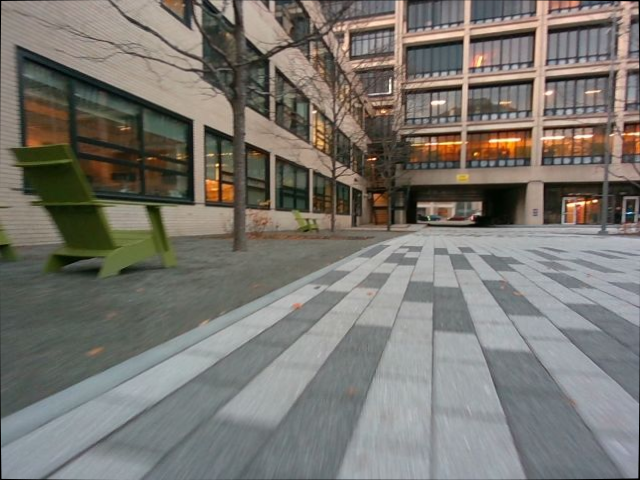}
      \caption{RGB input}
   	\end{subfigure}
  	\begin{subfigure}{0.28\columnwidth}
      \includegraphics[trim={0 0 0 0},clip, width=\columnwidth]{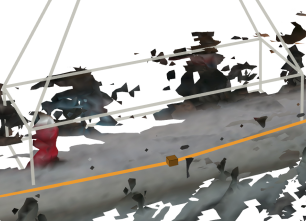}
      \includegraphics[trim={0 0 0 0},clip, width=\columnwidth]{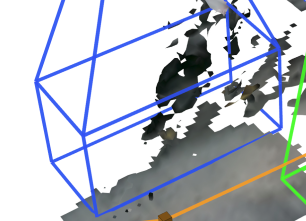}
      \includegraphics[trim={0 0 0 0},clip, width=\columnwidth]{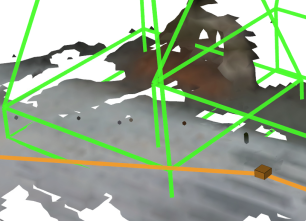}
      \caption{Khronos~\cite{Schmid24rss-khronos}}
   	\end{subfigure}
  	\begin{subfigure}{0.28\columnwidth}
      % {left  bottom  right  top}
      \includegraphics[trim={0 0 0 0},clip, width=\columnwidth]{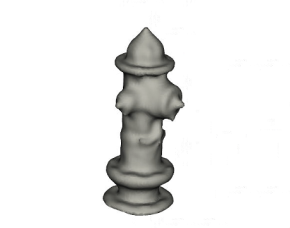}
      \includegraphics[trim={0 0 0 0},clip, width=\columnwidth]{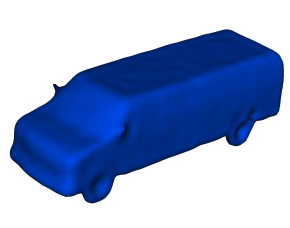}
      \includegraphics[trim={0 0 0 0},clip, width=\columnwidth]{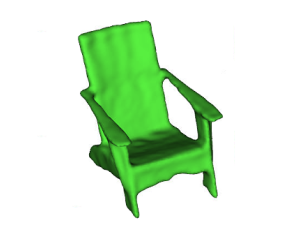}
      \caption{\methodname~(Ours)}
   	\end{subfigure}
    \caption{Object-level close-up results from the outdoor scene graph. From top to bottom: fire hydrant, car, and bench.
      (a)~RGB input image (\ie~$\imgRGB$ in \Cref{eq:shape_input}).
      (b)~Khronos~\cite{Schmid24rss-khronos}, an \RGBD-based method relying on instance-aware voxel clustering,
      fails to generate reliable object meshes due to sparse depth measurements in outdoor scenes,
      leading to under-segmentation with partial meshes.
      (c)~\methodname with hybrid \LiDAR-camera fusion successfully reconstructs fine-grained geometric details of each object via \CRISP~\cite{Shi25cvpr-CRISP}.}
    \label{fig:outdoor}
\end{figure}

As shown in \Cref{fig:mesh_comparison} and \Cref{fig:outdoor}, 
existing methods, such as Hydra~\cite{Hughes24ijrr-hydraFoundations} or Khronos~\cite{Schmid24rss-khronos}, suffer from degraded depth accuracy in outdoor scenes,
leading to incomplete scene meshes~(\Cref{fig:mesh_comparison}(a)) and under-segmented object reconstructions~(\Cref{fig:outdoor}(b)).
In contrast, our hybrid mode, using the ground-aware adaptive integration strategy presented in \Cref{sec:adaptive},
reduces mesh discontinuities and spurious holes in the scene reconstruction~(\Cref{fig:mesh_comparison}(c)),
while simultaneously recovering fine-grained object details via learned shape priors~(\Cref{fig:outdoor}(c)).

This complementary integration enables consistent metric scene meshes together with detailed,
instance-level object geometry, supporting our third claim on improved outdoor reconstruction quality.

\subsection{Quantitative and Qualitative Studies on RMCC}\label{sec:RMCC_ablation}

\begin{figure}[!t]
    \centering
    \begin{subfigure}{0.82\columnwidth}
      \includegraphics[trim={0 0 0 0},clip, width=0.065\textheight]{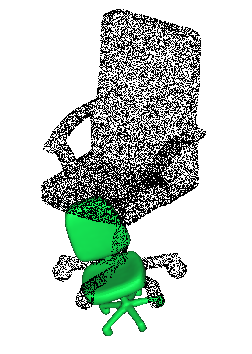}
      \includegraphics[trim={0 0 0 0},clip, width=0.065\textheight]{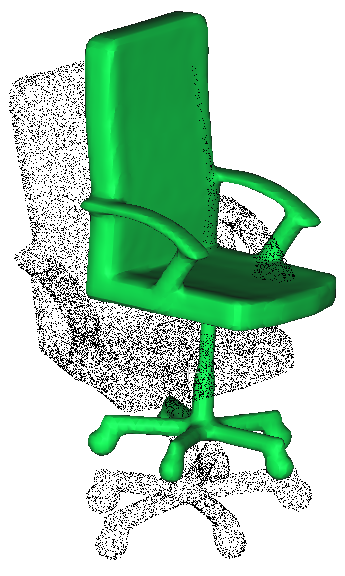}
      \includegraphics[trim={0 0 0 0},clip, width=0.065\textheight]{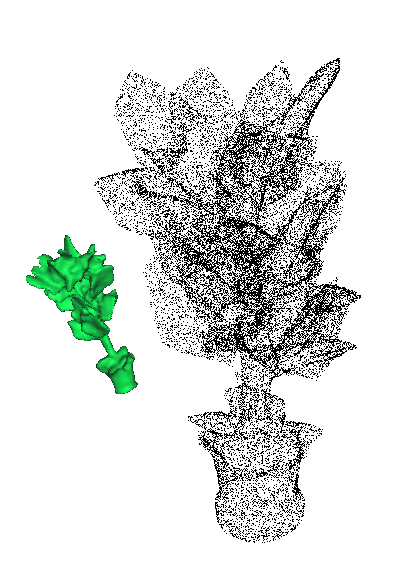}
      \includegraphics[trim={0 0 0 0},clip, width=0.065\textheight]{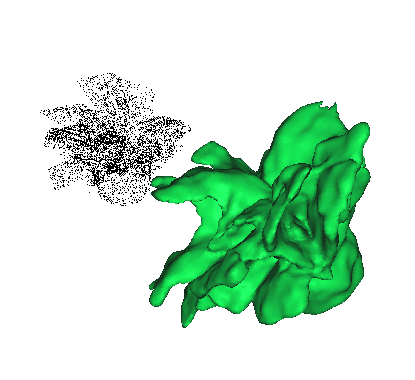}
    \caption{}
    \end{subfigure}
    \begin{subfigure}{0.82\columnwidth}
      \includegraphics[trim={0 0 0 0},clip, width=0.065\textheight]{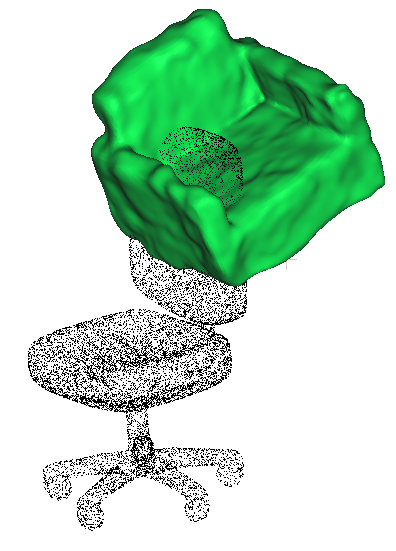}
      \includegraphics[trim={0 0 0 0},clip, width=0.065\textheight]{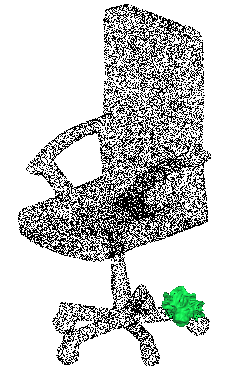}
      \includegraphics[trim={0 0 0 0},clip, width=0.065\textheight]{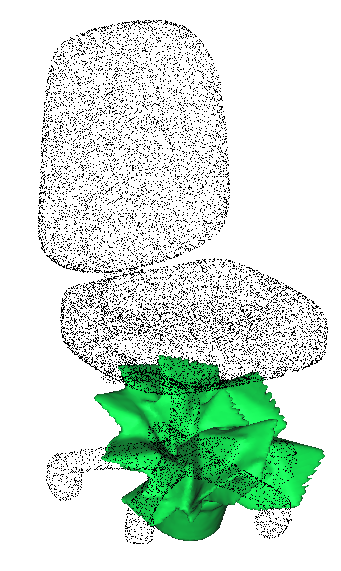}
      \includegraphics[trim={0 0 0 0},clip, width=0.065\textheight]{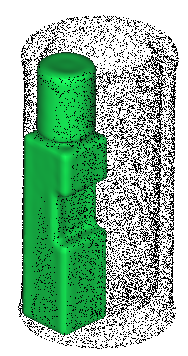}
    \caption{}
    \end{subfigure}
    \caption{
    Examples of erroneous shape estimates rejected by \RMCC.
    Green meshes show estimated shapes, and black points are sampled from ground truth.
    (a)~Intra-class errors with insufficient geometric overlap.
    (b)~Inter-class errors from class-agnostic \CRISP predictions.}
    \label{fig:overlap}
    \vspace{-3mm}
\end{figure}

In addition, we provide quantitative and qualitative results on the effect of our \RMCC.
Object shape prediction models are inherently vulnerable to partial masks
or degenerate viewpoints, which often lead to inaccurate shape estimates.
Because \CRISP operates as a class-agnostic shape estimator,
such ambiguous inputs can result in severe intra-class misalignments (\Cref{fig:overlap}(a))
or even generate shapes belonging to entirely different classes (\Cref{fig:overlap}(b)).
However, our \RMCC module effectively filters out these false cases by rejecting predictions
that exhibit insufficient image-space overlap according to the overlap score in \Cref{eq:overlapscore}.

\begin{figure}[!t]
    \centering
    \includegraphics[trim={0 0 0 0},clip, width=0.32\columnwidth]{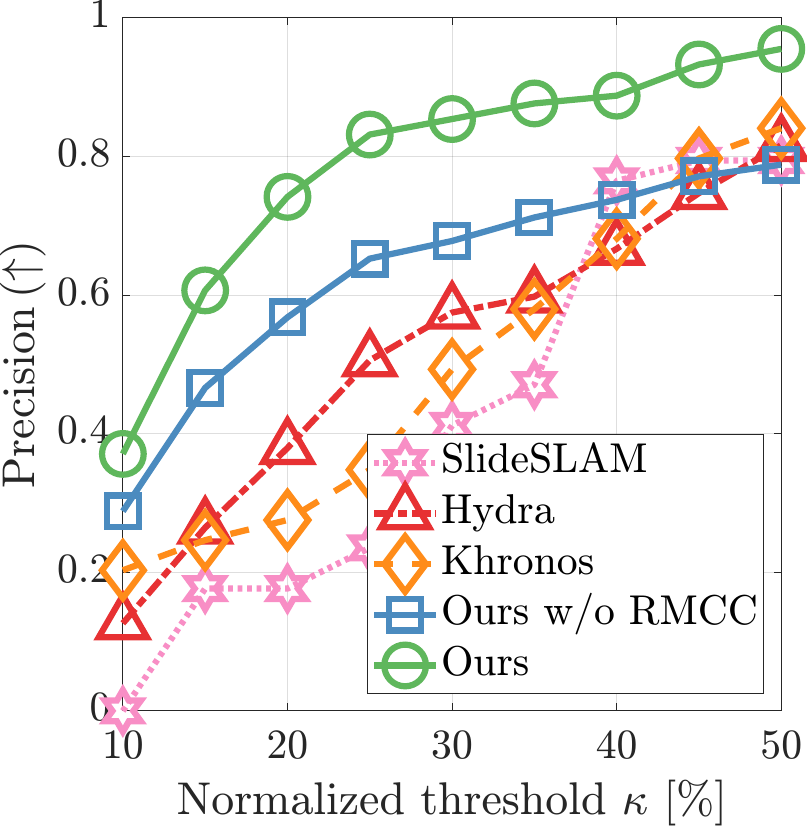}
    \includegraphics[trim={0 0 0 0},clip, width=0.32\columnwidth]{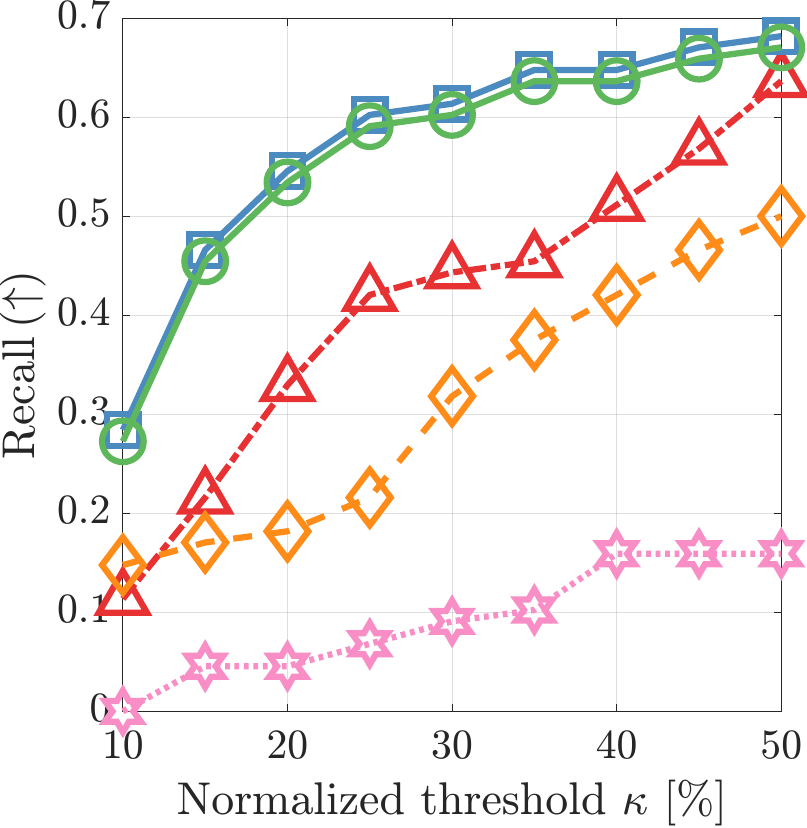}
    \includegraphics[trim={0 0 0 0},clip, width=0.32\columnwidth]{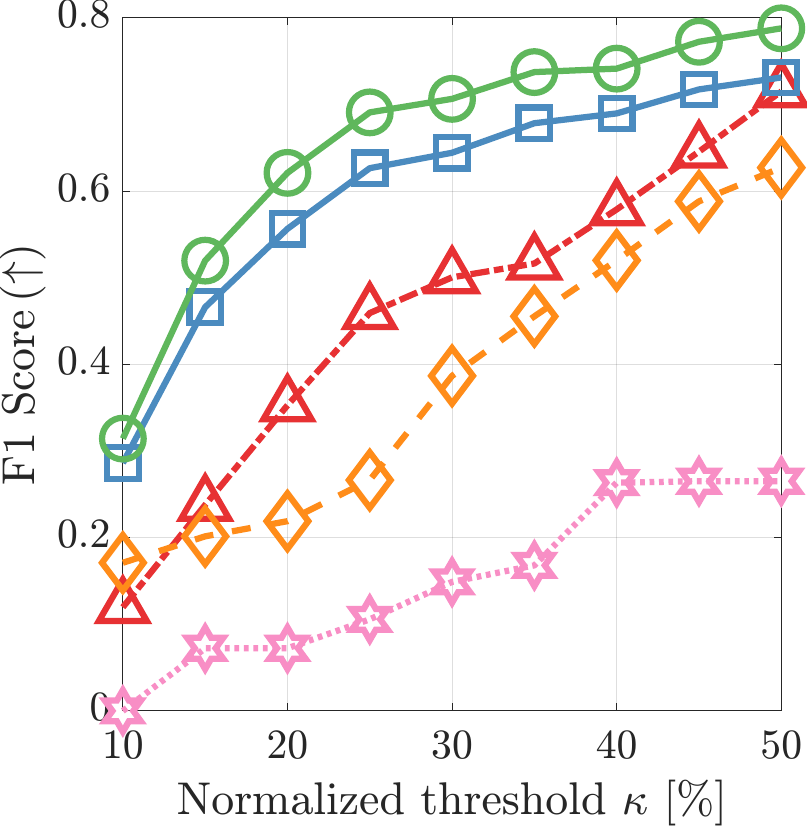}
    \caption{Precision, recall, and F$_1$ score as functions of the normalized threshold $\kappa$, applied as $\tau_d \leftarrow \kappa \cdot d_{\text{diag}}$,
    where $d_{\text{diag}}$ denotes the half-diagonal length of each object.}
    \label{fig:ratio}
\end{figure}

Furthermore, since \RMCC is driven by relative geometric proportions rather than absolute physical sizes,
we evaluate its performance using a relative threshold $\tau_d \leftarrow \kappa \cdot d_{\text{diag}}$,
where $d_{\text{diag}}$ is the half-diagonal length of the ground-truth object mesh.
As shown in \Cref{fig:ratio},
while \RMCC causes a slight drop in recall by rejecting some valid but uncertain estimates, 
enabling \RMCC substantially increases precision across all relative thresholds.
Consequently, \methodname with \RMCC achieves consistently higher F$_1$ scores
across the entire threshold range, confirming its efficacy as a robust post-verification step.

% As shown in \Cref{fig:ratio}, \RMCC consistently improved precision across all thresholds by filtering out degenerate shapes caused by partial or occluded observations.
% In contrast, recall was slightly lower with \RMCC enabled, since some valid but uncertain estimates were also rejected.
% However, this trade-off was favorable, as the precision gain outweighed the marginal recall loss,
% resulting in consistently higher F$_1$ scores for \methodname with \RMCC across the entire threshold range.
% Moreover, \RMCC substantially improved Chamfer distance, as shown in \Cref{fig:chamfer_distance},

% As shown in \Cref{fig:chamfer_distance}, \RMCC substantially improved class-wise Chamfer distance,
% indicating that spurious or erroneous shape predictions were successfully rejected while retaining reliable objects with high overlap scores (\ie \Cref{eq:overlapscore}).
% Overall, these results confirm that \RMCC served as an effective post-verification step that improved the reliability of object-level shape estimation.

\section{Conclusion and Lessons Learned} % \red{[0.5 column(s)]}
\label{sec:conclusions}
We have presented \textit{\methodname}, a scene graph-based system that integrates learned object-level pose and shape estimation within a hierarchical 3D scene graph framework.
In particular, we have demonstrated that a pose and shape estimation pipeline can be seamlessly integrated into the system,
providing improved object-level geometric representations while maintaining semantic grounding.
We have further demonstrated that fusing \LiDAR and camera information improves scene-level mesh completeness in outdoor environments.

Our experience also highlights the lack of outdoor benchmarks for object-level mesh reconstruction.
Existing public datasets commonly provide point-wise or pixel-wise annotations,
but rarely offer scene-scale, instance-level mesh ground truth from a mapping perspective.
This limits both training and evaluation of shape estimators such as \CRISP in real outdoor settings.
Future work will explore generative outdoor scene synthesis to create such benchmarks and improve generalization beyond controlled indoor environments.
% We will also study more diverse real-world deployments and failure cases involving noisy segmentation, heavy occlusion, and predictions with plausible silhouettes but incorrect 3D geometry.
% We further plan to explore integrating our mesh-based representation into multi-robot and multi-session scenarios,
% and applying the resulting rich geometric information to downstream tasks such as manipulation and human-robot interaction.

% Bibliography
% \clearpage
%{\footnotesize
\bibliographystyle{IEEEtran}
\bibliography{hydra_ref}
%}

\end{document}